\documentclass[10pt,twocolumn,letterpaper]{article}

\usepackage{conf}
\usepackage{algorithm}
\usepackage{algpseudocode}
\usepackage{times}
\usepackage{epsfig}
\usepackage[utf8]{inputenc} %
\usepackage[T1]{fontenc}    %
\usepackage{url}            %
\usepackage{booktabs}       %
\usepackage{amsfonts}       %
\usepackage{nicefrac}       %
\usepackage{microtype}      %
\usepackage[table]{xcolor}         %
\usepackage{amssymb}
\usepackage{amsmath}
\usepackage{tabularx}
\usepackage{multirow}
\usepackage{graphicx}
\usepackage{xcolor}
\usepackage{color}
\usepackage{caption}
\usepackage{subcaption}
\usepackage{enumitem}
\usepackage{array}
\usepackage{wasysym}

\usepackage{stfloats}

\newcommand{\method}{UniverSeg}

\usepackage{amsmath,amsfonts,bm}

\def\eqref#1{equation~\ref{#1}}

\def\1{\bm{1}}

\DeclareMathAlphabet{\mathsfit}{\encodingdefault}{\sfdefault}{m}{sl}
\SetMathAlphabet{\mathsfit}{bold}{\encodingdefault}{\sfdefault}{bx}{n}

\usepackage{array}
\newcolumntype{L}{>{\centering\arraybackslash}m{6cm}}
\newcolumntype{e}{>{\centering\arraybackslash}m{2cm}}

\newcommand{\subpara}[1]{\vspace{5pt} \noindent \textbf{#1}}

\usepackage[pagebackref=true,breaklinks=true,colorlinks,bookmarks=false]{hyperref}

\conffinalcopy %

\ifconffinal\fi

\usepackage[utf8]{inputenc} %
\usepackage[T1]{fontenc}    %
\usepackage{url}            %
\usepackage{booktabs}       %
\usepackage{amsfonts}       %
\usepackage{nicefrac}       %
\usepackage{microtype}      %
\usepackage{xcolor}         %
\usepackage{booktabs}
\usepackage{bm} %
\usepackage{balance}                %
\usepackage{setspace}
\usepackage{wrapfig}
\usepackage{multirow}
\usepackage{multicol}
\usepackage[per-mode=symbol]{siunitx} %
\usepackage{etoolbox}
\newtoggle{comments}

\definecolor{ultramarine}{RGB}{0,32,96}
\definecolor{firebrick}{RGB}{178,34,34}
\definecolor{navy}{RGB}{0,0,128}
\definecolor{forestgreen}{RGB}{0,128,0}

\newcommand{\authorcomment}[3]{{\color{#2}[\textbf{#1}:#3]}}
\newcommand{\draft}[1]{{\color{ultramarine}\itshape #1}}
\newcommand{\JJ}[1]{\authorcomment{J}{firebrick}{#1}}
\newcommand{\Victor}[1]{\authorcomment{Victor}{forestgreen}{#1}}
\newcommand{\Adrian}[1]{\authorcomment{Adrian}{navy}{#1}}
\newcommand{\John}[1]{\authorcomment{John}{ultramarine}{#1}}

\toggletrue{comments} %

\iftoggle{comments}{}{
    \renewcommand{\draft}[1]{}
    \renewcommand{\JJ}[1]{}
    \renewcommand{\Adrian}[1]{}
    \renewcommand{\Victor}[1]{}
    \renewcommand{\John}[1]{}
}

\begingroup
\lccode`\~=`\ %
\lowercase{%
  }%
\endgroup
\newcounter{word}

\begin{document}

   \abovedisplayskip=4pt
   \belowdisplayskip=4pt

\title{UniverSeg: Universal Medical Image Segmentation}

\author{Victor Ion Butoi\thanks{Denotes equal contribution}\\
MIT CSAIL\\
{\tt\small vbutoi@mit.edu}
\and
Jose Javier Gonzalez Ortiz\footnotemark[1]\\
MIT CSAIL\\
{\tt\small josejg@mit.edu}
\and
Tianyu Ma\\
Cornell University\\
{\tt\small tm544@cornell.edu}
\and
Mert R. Sabuncu\\
Cornell University\\
{\tt\small msabuncu@cornell.edu}
\and
John Guttag\\
MIT CSAIL\\
{\tt\small guttag@mit.edu}
\and
Adrian V. Dalca\\
MIT CSAIL \& MGH, HMS\\
{\tt\small adalca@mit.edu}
}

\maketitle
\ifconffinal\thispagestyle{empty}\fi

\begin{abstract}

\begin{figure*}[b]
  \centering
\includegraphics[width=\textwidth]{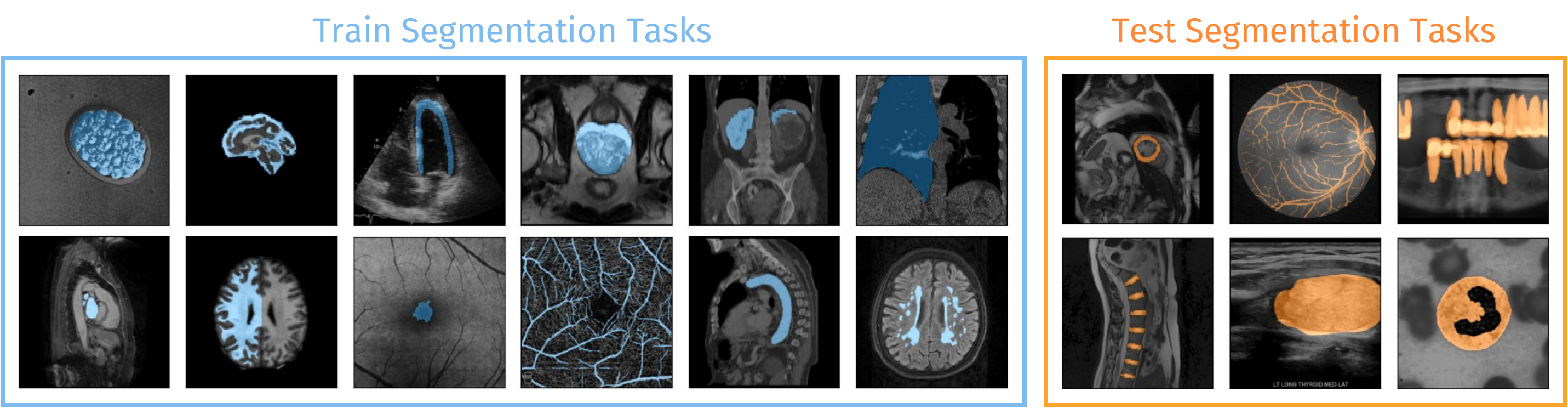}
  \caption{Medical segmentation involves many imaging types, biomedical domains, and target labels. We employ a large diverse set of training tasks \textbf{(blue)} to build a model that can segment unseen tasks \textbf{(orange)} without additional training.}
  \label{fig: teaser}
\end{figure*}

While deep learning models have become the predominant method for medical image segmentation, they are typically not capable of generalizing to unseen segmentation tasks involving new anatomies, image modalities, or labels.
Given a new segmentation task, researchers generally have to train or fine-tune models, which is time-consuming and poses a substantial barrier for clinical researchers, who often lack the resources and expertise to train neural networks.
We present UniverSeg, a method for solving unseen medical segmentation tasks without additional training.
Given a query image and example set of image-label pairs that define a new segmentation task, UniverSeg employs a new CrossBlock mechanism to produce accurate segmentation maps without the need for additional training. To achieve  generalization to new tasks, we have gathered and standardized a collection of 53 open-access medical segmentation datasets with over 22,000 scans, which we refer to as MegaMedical. We used this collection to train UniverSeg on a diverse set of anatomies and imaging modalities.
We demonstrate that UniverSeg substantially outperforms several related methods on unseen tasks, and thoroughly analyze and draw insights about important aspects of the proposed system. The \method{} source code and model weights are freely available at \texttt{\url{https://universeg.csail.mit.edu}}

\end{abstract}

\section{Introduction}
\label{sec:intro}

Image segmentation is a widely studied problem in computer vision and a central challenge in medical image analysis. Medical segmentation tasks can involve diverse imaging modalities, such as magnetic resonance imaging (MRI), X-ray, computerized tomography (CT), and microscopy; different biomedical domains, such as the abdomen, chest, brain, retina, or individual cells; and different labels within a region, such as heart valves or chambers (Figure~\ref{fig: teaser}). This diversity has inspired a wide array of segmentation tools, each usually tackling one task or a small set of closely related tasks~\cite{chen2021transunet, hyperDenseNet2019, huang2017densenet, isensee2021nnunet, ronneberger2015unet, sharma2010automated}. In recent years, deep-learning models have become the predominant strategy for medical image segmentation~\cite{kamnitsas2016deepmedic, milletari2016v, ronneberger2015unet}.

A key problem in image segmentation is \textit{domain shift}, where models often perform poorly given out-of-distribution examples. This is especially problematic in the medical domain where clinical researchers or other scientists are constantly defining new segmentation tasks driven by evolving populations, and scientific and clinical goals. To solve these problems they need to either train models from scratch or fine-tune existing models. Unfortunately, training neural networks requires machine learning expertise, computational resources, and human labor. This is infeasible for most clinical researchers or other scientists, who do not possess the expertise or resources to train models. In practice, this substantially slows scientific development. We therefore focus on avoiding the need to do \textit{any} training given a new segmentation tasks.

Fine-tuning models trained on the natural image domain can be unhelpful in the medical domain~\cite{raghu2019transfusion}, likely due to the differences in data sizes, features, and task specifications between domains, and importantly still requires substantial retraining. Some few-shot semantic segmentation approaches attempt to predict novel classes without fine-tuning in limited data regimes, but mostly focus on classification tasks, or segmentation of new classes within the same input domain, and do not generalize across anatomies or imaging modalities.

In this paper, we present \method{} -- an approach to learning a \textit{single} general medical-image segmentation model that performs well on a variety of tasks without any retraining, including tasks that are substantially different from those seen at training time. \method{} learns how to exploit an input set of labeled examples that specify the segmentation task, to segment a new biomedical image  in one forward pass. We make the following contributions.
\begin{figure*}[t]
\centerline{\includegraphics[width=0.95\textwidth]{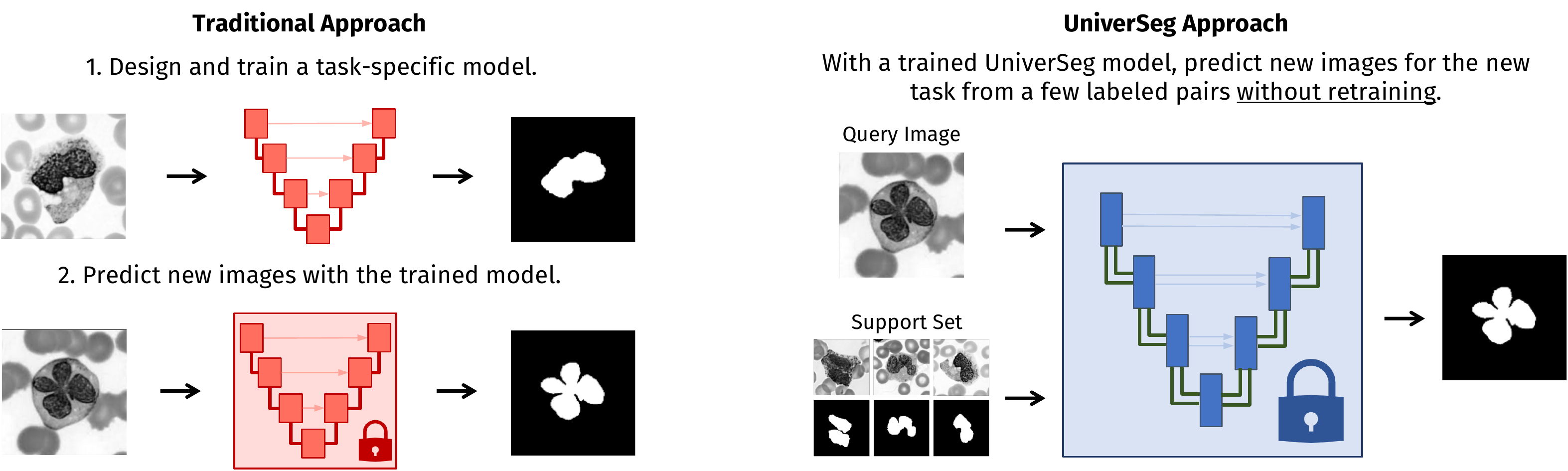}}
\vspace{-3pt}
\caption{
\textbf{Workflow for inference on a new task, from an unseen dataset.} Given a new task, traditional models \textbf{(left)} are trained before making predictions. \method{} \textbf{(right)} employs a \textit{single} trained model which can make predictions for images (queries) from the new task with a few labeled examples as input (support set), without additional fine-tuning. \vspace{-8pt}
}
\label{fig: paradigm}
\end{figure*}

\vspace{-3pt}
\begin{itemize}[itemsep=-2pt]
    \item We propose \method{} -- a framework that enables solving new segmentation tasks without retraining, using a novel flexible CrossBlock mechanism that transfers information from the example set to the new image.
    \item We demonstrate that \method{} substantially outperforms several models across diverse held-out segmentation tasks involving unseen anatomies, and even approaches the performance of fully-supervised networks  trained specifically for those tasks.
    \item In extensive analysis, we show that the generalization capabilities of \method{} are linked to task diversity during training and image diversity during inference.
\end{itemize}

\method{} source code and model weights are available at \texttt{\url{https://universeg.csail.mit.edu}}

\vspace{-3pt}
\section{Related Works}
\vspace{-2pt}
\subpara{Medical Image Segmentation.}
Medical image segmentation has been widely studied, with state-of-the-art methods training convolutional neural networks in a supervised fashion, predicting a label map for a given input image~\cite{hyperDenseNet2019, huang2017densenet, isensee2021nnunet, kamnitsas2017efficient, ronneberger2015unet}. For a new segmentation problem, models are typically trained from scratch, requiring substantial design and tuning. 

Recent strategies, such as the nnUNet~\cite{isensee2021nnunet}, automate some design decisions such as data processing or model architecture, but still incur substantial overhead from training. In contrast to these methods, \method{} generalizes to new medical segmentation tasks without training or fine-tuning.

\subpara{Multi-task Learning.}
Multi-Task Learning (MTL) frameworks learn several tasks simultaneously~\cite{caruana1997multitask, evgeniou2004regularized, sener2018multi}. For medical imaging, this can involve multiple modalities~\cite{moeskops2016deep}, population centers~\cite{liu2020ms}, or anatomies~\cite{navarro2019shape}. However, the tasks are always pre-determined by design: once trained, each network can only solve tasks presented during training. \method{} overcomes this limitation, enabling tasks to be dynamically specified during inference.

\begin{figure*}[t]
\centerline{\includegraphics[width=\textwidth]{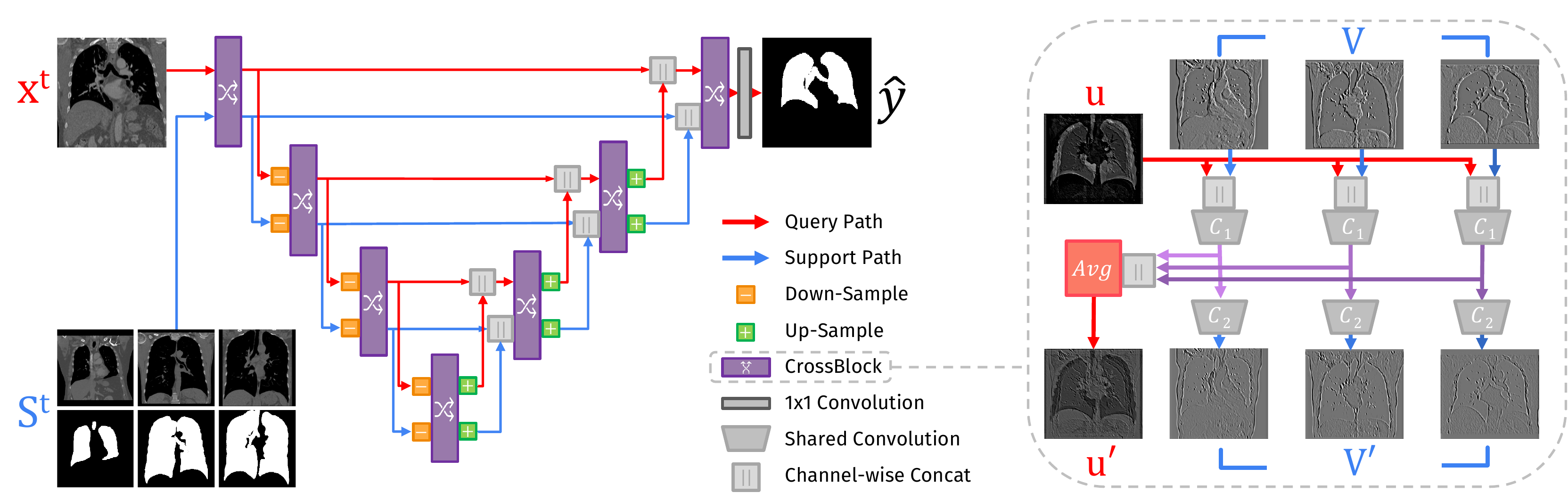}}
\vspace{-3pt}
\caption{A \method{} network \textbf{(left)} takes as input a query image and a support set of image and label-maps (pairwise concatenated in the channel dimension) and employs multi-scale CrossBlock features. A CrossBlock \textbf{(right)} takes as input representations of the query $u$ and support set $V=\{v_i\}$, and interacts $u$ with each support entry $v_i$ to produce $u^\prime$ and $V^\prime$.
\vspace{-4pt}}
\label{fig: Method-Diagram}
\end{figure*}

\subpara{Transfer Learning.}
Transfer learning strategies involve fine-tuning pre-trained models, often from a different domain~\cite{long2016unsupervised, sun2016return}. This is used in medical image segmentation starting with models trained on natural images~\cite{novelTransfer2021, ghafoorian2017transfer, jiang2018retinal, zhao2019data, zhou2021models}, where the amount of data far exceeds the amount in the target biomedical domain. However, this technique still involves substantial training for each new task, which \method{} avoids. Additionally, the differences between medical and natural images often make transfer learning from large pre-trained models unhelpful~\cite{raghu2019transfusion}.

\subpara{Optimization-based Meta-Learning.}
Optimization-based meta-learning techniques often learn representations that minimize downstream fine-tuning steps by using a few examples per task, sometimes referred to as few-shot learning~\cite{finn2017model, nichol2018reptile, vinyals2016matching, snell2017prototypical}. Meta-learning via fine-tuning has been studied in medical image segmentation to handle multiple image modalities~\cite{zhang2021modality}, anatomies~\cite{zhang2021domain}, and generalization to different targets~\cite{khadka2022meta, khandelwal2020domain, singh2021metamed}. While these strategies reduce the amount of data and training required for downstream tasks~\cite{he2019rethinking}, fine-tuning these models nevertheless requires machine learning expertise and computational resources, which are often not available to medical researchers.

\subpara{Few-shot Semantic Segmentation.}
Few-shot (FS) methods adapt to new tasks from few training examples, often by fine-tuning pretrained networks~\cite{finn2017model, nichol2018reptile, vinyals2016matching, snell2017prototypical}. Some few-shot semantic segmentation models generate predictions for new images (queries) containing unseen classes from just a few labeled examples (support) without additional retraining. One strategy prevalent in both natural image~\cite{fwbNguyen2019, seo2022task, zhang2019canet} and medical image~\cite{ding2023few, li2022prototypical, pandey2022robust, shen2022q} FS segmentation methods is to employ large pre-trained models to extract deep features from the query and support images. These methods often involve learning meaningful prototypical representations for each label~\cite{panet2019Wang}. Another medical FS segmentation strategy uses self-supervised learning to make up for the lack of training data and tasks~\cite{hansen2022anomaly, ouyang2020alpnet}. In contrast to \method{}, these methods, focused on limited data regimes, tackle specific tasks involving generalizing to new classes in a particular subdomain, like abdominal CT or MRI scans~\cite{hansen2022anomaly, ouyang2020alpnet, roy2020SEnet, tang2021recurrent}.

In our work, we focus on avoiding \textit{any} fine-tuning, even when given many examples for a new task, to avoid requiring the clinical or scientific user to have machine learning expertise and compute resources. Our proposed framework draws inspiration from ideas from some few-shot learning solutions, but aims to generalize to a universally broad set of anatomies, modalities, and datasets -- even those completely unseen during training.

\vspace{-3pt}
\section{\method{} Method}
\vspace{-3pt}

Let~$t$ be a segmentation task comprised of a set of image-label pairs~$\{(x^{t}_i, y^{t}_i)\}^N_{i=1}$.
Common segmentation strategies learn parametric functions~$\hat{y} = f^{t}_\theta(x)$, where~$f^{t}_\theta$ is most often modeled using a convolutional neural network that estimates a label map~$\hat{y}$ given an input image~$x$. By construction,~$f^{t}_\theta$ only learns to predict segmentations for task~$t$.

In contrast, we learn a universal function~\mbox{$\hat{y} = f_\theta(x^{t}, S^{t})$} that predicts a label map for input~$x^{t}$ of task~$t$, according to the task-specifying support~$S^{t} = \{(x_j^{t}, y_j^{t})\}_{j=1}^n$ comprised of example image-label pairs available for $t$.

\subsection{Model}

We implement~$f_\theta$ using a fully convolutional neural network illustrated in Figure~\ref{fig: Method-Diagram}. We first introduce the proposed building blocks: the \emph{cross-convolution} layer and the \mbox{CrossBlock} module. We then specify how we combine these blocks into a complete segmentation network.

\subpara{CrossBlock.} To transfer information between the support set and query image, we introduce a \emph{cross-convolution} layer that interacts a query feature map $u$ with a set of support feature maps $V = \{v_i \}_{i=1}^n$:
\begin{equation}
\begin{aligned}
&\text{CrossConv}(u, V; \theta_z) = \{z_i\}_{i=1}^n,\\
&\qquad\text{for}\; z_i = \text{Conv}(u || v_i; \theta_z),
\end{aligned}
\end{equation}
where $||$ is the concatenation operation along the feature dimension and Conv$(x; \theta_z)$ is a convolutional layer with learnable parameters $\theta_z$. Due to the weight reuse of~$\theta_z$, cross-convolution operations are permutation invariant with respect to $V$.
From this layer, we design a higher-level building block that produces updated versions of query representation $u$ and support $V$ at each step in the network:
\begin{align}
&\text{CrossBlock}(u, V; \theta_z, \theta_v) = (u^\prime, V^\prime), \text{where:} \\
&\qquad z_i = A(\text{CrossConv}(u, v_i; \theta_z)) \quad \text{for}\; i = 1,2,\ldots ,n \nonumber\\
&\qquad u^\prime = 1/n \textstyle\sum_{i=1}^n z_i \nonumber\\
&\qquad v^\prime_i = A(\text{Conv}(z_i; \theta_v)) \quad \text{for}\; i = 1,2,\ldots ,n, \nonumber 
\end{align}
where $A(x)$ is a non-linear activation function.
This strategy enables the representations of each support set entry and query to interact with the others through their average representation, and facilitates variably sized support sets.

\subpara{Network.} To integrate information across spatial scales, we compose the CrossBlock modules in an encoder-decoder structure with residual connections, similarly to the popular UNet architecture (Figure~\ref{fig: Method-Diagram}). The network takes as input the query image $x^{t}$ and support set~$S^{t}=\{(x_i^{t},y_i^{t})\}_{i=1}^n$ of image and label-map pairs, each concatenated channel-wise, and outputs the segmentation prediction map $\hat{y}^{t}$.

Each level in the encoder path consists of a CrossBlock followed by a spatial down-sampling operation of both query and support set representations. Each level in the expansive path consists of up-sampling both representations, which double their spatial resolutions, concatenating them with the equivalently-sized representation in the encoding path, followed by a CrossBlock. We perform a single 1x1 convolution to map the final query representation to a prediction.

\subsection{Training}

Algorithm~\ref{alg:training-loop} describes \method{} training using a large and varied set of training tasks~$\mathcal{T}$ and the loss
\begin{equation}
    \mathcal{L}(\theta; \mathcal{T})
= \mathbb{E}_{t \in \mathcal{T}} \mathbb{E}_{(x^{t}, y^{t}), S^{t}}  \Big[ \mathcal{L}_\textrm{seg}(f_\theta(x^{t}, S^{t}), y^{t}) \Big],
\end{equation}
where $x^{t} \notin S^{t}$, and~$\mathcal{L}_\textrm{seg}(\hat{y}, y^{t})$ is a standard segmentation loss like cross-entropy or soft Dice~\cite{milletari2016v}, capturing the agreement between the predicted~$\hat{y}$ and ground truth~$y_t$.

\subpara{Data Augmentation.}
 We employ data augmentation to grow the diversity of training tasks and increase the number of effective training examples belonging to any particular task.

\textit{In-Task Augmentation ($\text{Aug}_t(x, y)$).}
To reduce overfitting to individual subjects, we perform standard data augmentation operations, like affine transformations, elastic deformation, or adding image noise to the query image and \textit{each entry} of the support set independently.

\textit{Task Augmentation ($\text{Aug}_T(x,y,S)$).}
Similar to standard data augmentation that reduces overfitting to training examples, augmenting the training \textit{tasks} is useful for generalizing to \textit{new tasks}, especially those far from the training task distribution. We introduce task augmentation -- alterations that modify all query and support images, and/or all segmentation maps, with the same type of task-changing transformation. %
Example task augmentations include edge detection of the segmentation maps or a horizontal flip to all images and labels. We provide a list of all augmentations and the parameters we used in the supplemental Section~\ref{sec: supp-aug}.

\begin{algorithm}[t]
\caption{\method{} Training Loop using SGD with learning rate~$\eta$ over tasks~$\mathcal{T}$, main architecture~$f_\theta$, in-task augmentations~$\text{Aug}_t$ and task augmentations~$\text{Aug}_T$}
\label{alg:training-loop}
\begin{algorithmic}
\For{$k = 1,\ldots, \text{NumTrainSteps}$}
\State $t\sim \mathcal{T}$ \Comment{Sample Task}
\State $(x_i^{t}, y_i^{t}) \sim t$ \Comment{Sample Query}
\State $S^{t} \gets \{ (x_j^{t}, y_j^{t})\}_{j\neq i}^n$ \Comment{Sample Support}
\State $x_i^{t}, y_i^{t} \gets \text{Aug}_t(x_i^{t}, y_i^{t})$ \Comment{Augment Query}
\State $S^{t} \gets \{ \text{Aug}_t(x_j^{t}, y_j^{t}) \}_j^n$ \Comment{Augment Support}
\State $x_i^{t}, y_i^{t}, S^{t} \gets \text{Aug}_T(x_i^t, y_i^{t}, S^{t})$ \Comment{Task Aug}
\State $\hat{y}_i \gets f_\theta(x_i^{t}, S^{t})$ \Comment{Predict label map}
\State $\ell \gets \mathcal{L}_{\text{seg}}(\hat{y}_i, {y}_i^{t})$ \Comment{Compute loss}
\State $\theta \gets \theta - \eta \nabla_\theta \ell$ \Comment{Gradient step}
\EndFor
\end{algorithmic}
\end{algorithm}

\subsection{Inference}

For a given query image $x^t$, \method{} predicts segmentation $\hat{y}=f_\theta(x^t,S^t)$ given a support set $S^{t}$ 
, where the prediction quality depends on the choice of the support set $S^{t}$.
To reduce this dependence, and to take advantage of more data when memory constraints limit the support set size at inference, we combine predictions from an ensemble of $K$ independently sampled support sets $\{S^{t}_{i}\}_{i=1}^K$
 as their the pixel-wise average to produce the prediction
 $
     \hat{y} = \frac{1}{K} \sum_{k=1}^K f_\theta(x, S^{t}_k).
 $

\section{MegaMedical Dataset}

To train our universal model~$f_\theta$, we employ a set of segmentation tasks that is large and diverse, so that it is able to generalize to new tasks. We compiled MegaMedical -- an extensive collection of open-access medical segmentation datasets with diverse anatomies, imaging modalities, and labels. It is constructed from 53 datasets encompassing 26 medical domains and 16 imaging modalities.

We standardize data across the wildly diverse formats of original datasets, processed images, and label maps.
We also expand the training data using synthetic segmentation tasks to further increase the training task diversity.
Because of individual dataset agreements, we are prohibited from re-releasing our processed version of the datasets. Instead, we will provide data processing code to construct MegaMedical from its source datasets.

\subpara{Datasets.}
MegaMedical features a wide array of biomedical domains, such as eyes~\cite{STARE, OCTA500, Rose, IDRID, DRIVE}, lungs~\cite{CheXplanation, LUNA, MSD}, spine vertebrae~\cite{SpineWeb}, white blood cells~\cite{WBC}, abdominal~\cite{LiTS, NCI-ISBI, KiTS, AMOS, CHAOSdata2019, SegTHOR, BTCV, I2CVB, Promise12, Word, AbdomenCT-1K, SCD, MSD}, and brain~\cite{BRATS, MCIC, ISLES, WMH, BrainDevelopment, OASIS-data, PPMI, LGGFlair, MSD}, among others.
Supplemental Table~\ref{tab: datasets} provides a detailed list of MegaMedical datasets.
Acquisition details, subject age ranges, and health conditions are different for each dataset.
We provide preprocessing and data normalization details in supplemental Section~\ref{sec: supp-megamedical}.

\subpara{Medical Image Task Creation.}
While datasets in MegaMedical feature a variety of imaging tasks and label protocols, in this work we focus on the general problem of 2D binary segmentation.
For datasets featuring 3D data, for each subject, we extract the 2D mid-slice of the volume along all the major axes.
When multiple modalities are present, we include each modality as a new task.
For datasets containing multiple segmentation labels, we create as many binary segmentation tasks as available labels.
All images are resized to $128 \times 128$ pixels and intensities are normalized to the range [0,1].

\subpara{Synthetic Task Generation.} We adapt the image generation procedure involving random synthetic shapes described in SynthMorph~\cite{hoffmann2022synthmorph} to produce a thousand synthetic tasks to be used alongside the medical tasks during training. We detail the generation process and include examples of synthetic tasks in supplemental Section~\ref{sec: supp-synth}.

\section{Experiments}
\label{Experiments}

We start by describing experimental details. The first set of experiments compares the performance of \method{} in the held-out datasets against several single-pass methods used in few-shot learning. We then report on a variety of analyses, including ablations of modeling decisions, and the effect of training task diversity, support set size, and number of examples available for a new task.

\subsection{Experimental Setup}

\subpara{Model}.
We implement the network in UniverSeg (Figure~\ref{fig: Method-Diagram}) using an encoder with 5 CrossBlock stages and a decoder with 4 stages, with 64 output features per stage and LeakyReLU non-linearities after each convolution. We use bilinear interpolation when downsampling or upsampling.

\subpara{Data}.
For each dataset $d$, we construct three disjoint splits $d = \{d_\text{support}, d_\text{dev}, d_\text{test}\}$ with 60\%, 20\%, and 20\% of the subjects, respectively. 
Similar to dataset generalization~\cite{triantafillou2021learning}, we divide the available datasets into a training set $\mathcal{D}^T$ and a held-out test set $\mathcal{D}^H$.
We train models using the support and development splits of the training datasets $\{ d_\text{support} | d \in \mathcal{D}^T\}$. We performed model selection and hyper-parameter tuning using the development split of held-out dataset WBC, and trained models until they stopped improving in the $d_\text{dev}$ split, averaged across the held-out datasets.
We report results using the unseen test split of the held-out datasets $\{ d_\text{test} | d \in \mathcal{D}^H\}$.
Support set image-label pairs are sampled with replacement from each dataset's support split.

For held-out datasets, we evaluated three datasets containing anatomies represented in the training datasets (ACDC~\cite{ACDC} and SCD~\cite{SCD} (heart), and STARE\cite{STARE} (retinal blood vessels)), and three datasets of anatomies not covered by the rest of MegaMedical (PanDental~\cite{PanDental} (mandible), SpineWeb~\cite{SpineWeb} (vertebrae), and WBC~\cite{WBC} (white blood cells). 

\subpara{Few-Shot Baselines}.
We compare \method{} models to three segmentation methods from the few-shot (FS) literature, since these approaches also predict the segmentation of a query image given a support set of image-label pairs, although they were designed for the low-data regime.
SE-net~\cite{roy2020SEnet} features a fully-convolutional network, squeeze-excitation blocks, and a UNet-like model architecture.
ALPNet~\cite{ouyang2020alpnet} and PANet~\cite{panet2019Wang}, employ prototypical networks that extract prototypes from their inputs to match the given query with the support set. While ALPNet also employs a self-supervised method to generate additional label maps in settings with few tasks, we omit this step since MegaMedical includes a large collection of tasks.

Unlike \method{}, these methods were designed to generalize to similar tasks, such as different labels in the same anatomy and image type, or different modalities for the same anatomy. To make the comparison to \method{} fair, we make several additions to the training and inference procedures of these baselines as described below, and chose the best performing variant of each baseline.

\subpara{Supervised Task-Specific Models}.
While it is often impractical for clinical researchers to train individual networks for each task, for evaluation we train a set of task-specific networks to serve as an upper bound of supervised performance on the held-out datasets.
We employ the widely-used nnUNet~\cite{isensee2021nnunet}, which automatically configures the model and training pipeline based on data properties. Each model is task-specific, using the support and development splits for training and model selection, respectively. We report results on the test split.

\subpara{Evaluation}.
We evaluate models on the held-out datasets $\mathcal{D}^H$ using the test split for query images and the support split for support-sets.
For all methods, unless specified otherwise, we perform 5 independent predictions per test subject using randomly drawn support sets, and ensemble the predictions.
We enforce that the same random support sets are used for all methods.
We evaluate predictions using the Dice score~\cite{dice1945measures} (0 - 100, 0=no overlap, 100=perfect match), which quantifies the region overlap between two regions and is widely used in medical segmentation.
For tasks with more than one label, we average Dice across all labels.
For datasets with multiple tasks, we average performance across all tasks.
We estimate prediction variability using subject bootstrapping, with 1,000 independent repetitions. At each repetition, we treat each task independently, sampling subjects with replacement, and report the standard deviation across bootstrapped estimates.

\begin{figure*}[t]
\centerline{\includegraphics[width=\textwidth]{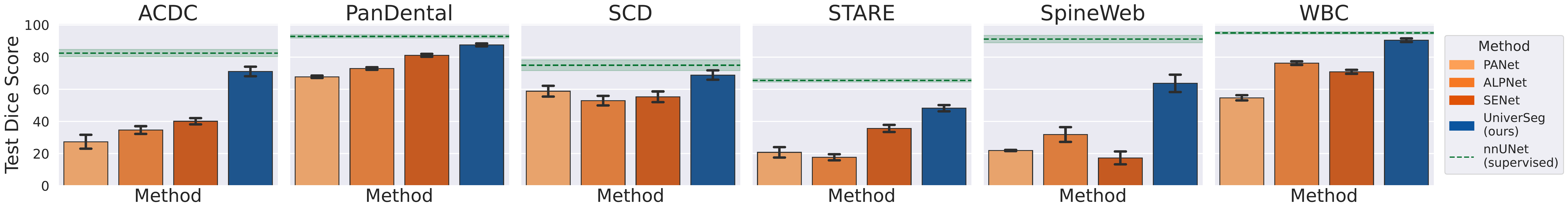}}
\caption{\textbf{Average Dice score per each held out dataset}. Performance of \method{} and several few-shot baselines, and the upper bound of each dataset determined by the individual fully-trained networks. For each of the unseen datasets, we average across tasks and subjects, and show the bootstrap variability in the error bars.
}
\label{fig: main-results}
\vspace{3pt}
\centerline{\includegraphics[width=0.9\textwidth]{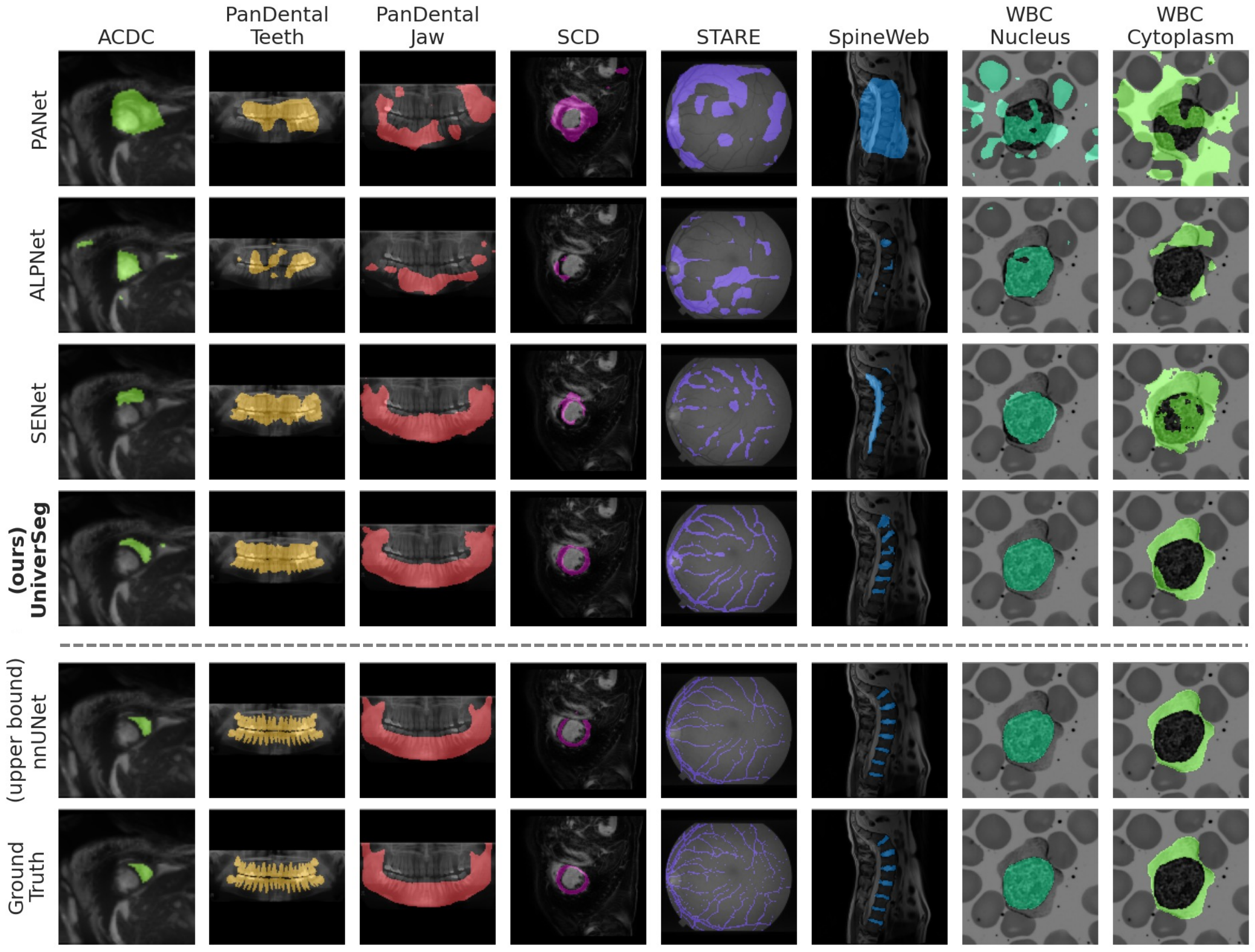}}
\caption{\textbf{Example model predictions for unseen tasks}. For a randomly sampled image per held-out task, we visualize the predictions of \method{}, few-shot baselines, and individually trained nnUNet models, along with ground truth maps. \vspace{-2pt}}
\label{fig: main-examples}
\end{figure*}

\subpara{Training}.
We train networks with the Adam optimizer~\cite{kingma2014adam} 
and soft Dice loss~\cite{milletari2016v, sudre2017generalised}.
For the ALPNet and PANet baselines, we add an additional prototypical loss term as described in their original works.
Models trained with cross-entropy performed substantially worse than soft Dice.

While the original baseline methods were not introduced with significant data augmentation, we trained all \method{} and FS models with and without the proposed augmentation transformations, and report results on the best-performing setting.
Unless specified otherwise, models are trained using a support size of 64. While the baselines were originally designed with small support sizes (1 or 5) as they tackled the few-shot setting, we found that training and evaluating them with larger support sizes improved their performance.

\subpara{Implementation}. We provide additional implementation and experimental details in supplemental Section~\ref{sec:impl}.
Code and pre-trained model weights for \method{} are available at \texttt{\url{https://universeg.csail.mit.edu}}.

\begin{table}
\resizebox{\columnwidth}{!}{%
\begin{tabular}{lrrr}
     \textbf{Model} & \textbf{\#Params} & \textbf{Runtime ms} & \textbf{Dice Score} \\
\toprule
           PANet &         14.71 &      240.0 $\pm$ 1.8 &       41.8 $\pm$ 1.3 \\
          ALPNet &         43.02 &      527.7 $\pm$ 8.7 &       47.8 $\pm$ 1.1 \\
           SENet &          0.92 &        4.1 $\pm$ 0.8 &       50.1 $\pm$ 1.3 \\
UniverSeg (ours) &          1.18 &      142.0 $\pm$ 0.4 & \textbf{71.8 $\pm$ 0.9} \\
   \midrule
   nnUNet (sup.) &     17${\times}$ 1.87  &      17${\times}$ 1.4${\cdot}10^7$ &       84.4 $\pm$ 1.0 \\
\bottomrule
\end{tabular}

}
\caption{
\textbf{Performance Summary}.
For \method{} and each FS baseline we report model size (in millions), inference run-time, and average held-out Dice score (with bootstrapping standard deviation) .
As an upper bound, we include the set of 17 individually trained task-specific nnUNets for the 6 held-out datasets, where their run-time is their cumulative required training time.
}
\label{tab:main-results}
\end{table}

\subsection{Task Generalization Results}

First, we compare the segmentation quality of \method{} with FS baselines and the task-specific upper bounds.
Our primary goal is to assess the effectiveness of \method{} in solving tasks from unseen datasets. Figure~\ref{fig: main-results} presents the average Dice scores per dataset for each method, and Figure~\ref{fig: main-examples} presents example segmentation results for each method and dataset.

\subpara{Few-shot methods}. \method{} significantly outperforms all FS methods in all held-out datasets. %
For each FS method, we report the best-performing model, which involved adding components of the \method{} training pipeline.
In the supplemental material, we show that few-shot methods perform worse when trained with a support set size of 1 and without ensembling, as they were originally introduced.  

\method{} outperforms the highest performing baseline for all datasets with Dice improvements ranging from 7.3 to 34.9.
Figure~\ref{fig: main-examples} also shows clear qualitative improvements in the predicted segmentations.
Given the similarities between SENet and \method{} (fully convolutional UNet-like structure), these results suggest that the proposed CrossBlock is better suited to transferring spatial information from the support set to the dquery.
Table~\ref{tab:main-results} shows that \method{} also requires fewer model parameters than PANet, ALPNet, and the nnUNets, and a similar number to SENet.

\subpara{Task-specific networks}.
For some datasets like PanDental or WBC, \method{} performs competitively with the supervised task-specific networks, which were extensively trained on each of the held-out tasks, and are unfeasible to run in many clinical research settings.
Moreover, from the qualitative results of Figure~\ref{fig: main-examples}, we observe that segmentations produced by \method{} more closely match those of the supervised baselines than those of any other few-shot segmentation task, especially in challenging datasets like SpineWeb or STARE.

\begin{figure}[t]
\centerline{\includegraphics[width=\columnwidth]{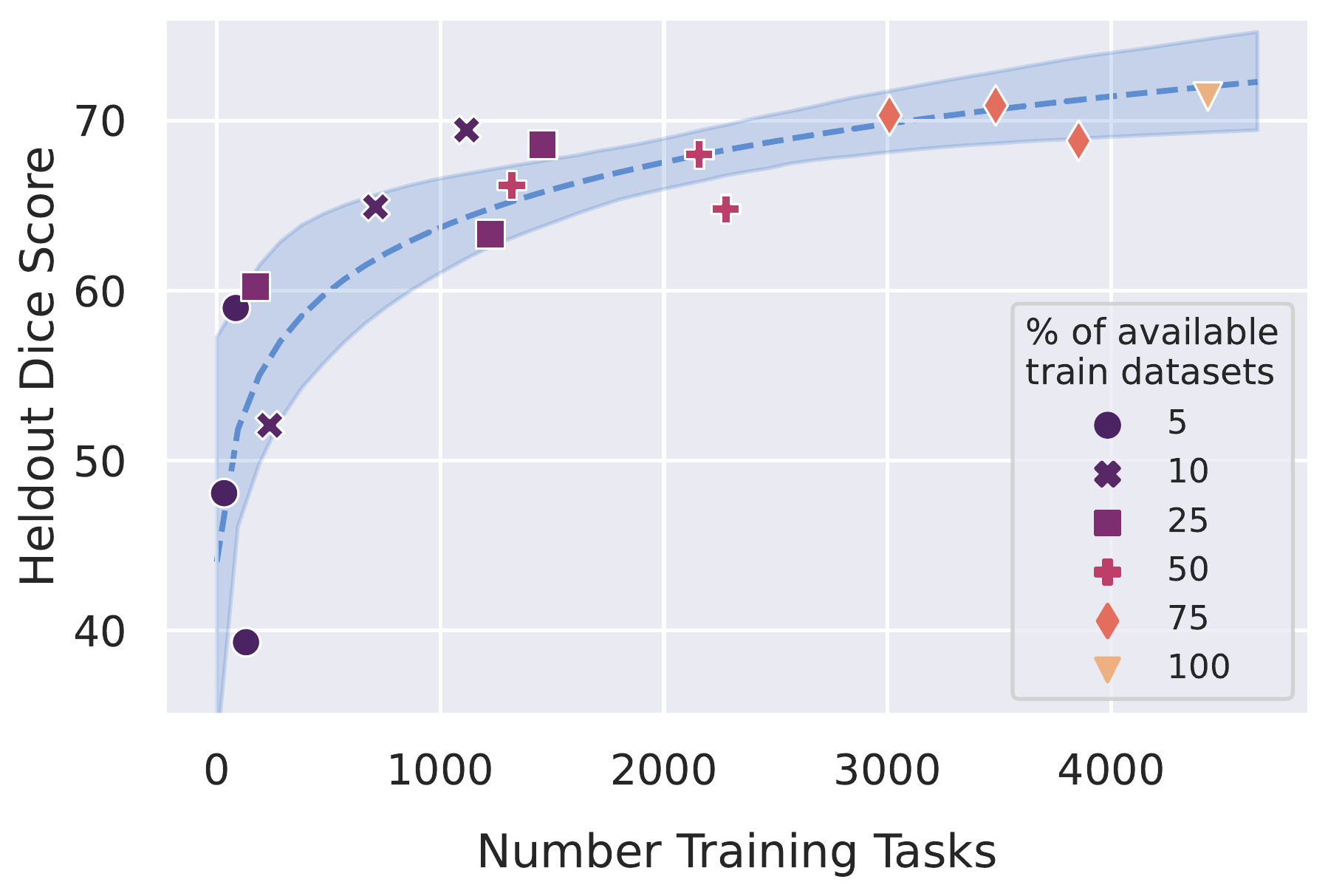}}
\caption{\textbf{Average held-out Dice versus the number of training tasks}.
Points represent individual \method{} networks trained on a percentage of available training datasets and shown in terms of the number of underlying training tasks. In blue, we report a logarithmic fit to the data and 95\% confidence intervals obtained by bootstrapped fits. 
}
\label{fig: TrainTasksPercent}
\end{figure}

\subsection{Analysis}

We analyze how several of the data, model, and training decisions affect the performance of \method{}.

\subpara{Task Quantity and Diversity.}
We study the effect of the number of datasets and individual tasks used for training \method{}.
We leave out synthetic tasks for this experiment, and train models on random subsets of the MegaMedical training datasets.

Figure~\ref{fig: TrainTasksPercent} presents performance on the held-out datasets for different random subsets of training datasets.
We find that having more training tasks improves the performance on held-out tasks. 
In some scenarios, the \textit{choice} of datasets has a substantial effect.
For instance, for models trained with $10\%$ of the datasets, the best model outperforms the worst one by  17.3 Dice points, and comparing those subsets we find that the best performing one was trained on a broad set of anatomies including heart, abdomen, brain, and eyes; while the least accurate model was trained on less common lesion tasks, leading to worse generalization.

\begin{table}
\centering
\rowcolors{2}{white}{gray!15}
\resizebox{0.85\columnwidth}{!}{%
\begin{tabular}{ccccr}
\textbf{Synth} & \textbf{Medical} & \textbf{In-Task} & \textbf{Task} & \textbf{Dice Score} \\
\toprule
          $\checkmark$ &               &               &            &       61.7 $\pm$ 1.5 \\
            &             $\checkmark$ &               &            &       62.7 $\pm$ 1.1 \\
          $\checkmark$ &             $\checkmark$ &               &            &       64.5 $\pm$ 1.0 \\
            &             $\checkmark$ &             $\checkmark$ &            &       67.0 $\pm$ 0.9 \\
            &             $\checkmark$ &               &          $\checkmark$ &       70.4 $\pm$ 1.3 \\
            &             $\checkmark$ &             $\checkmark$ &          $\checkmark$ &       70.0 $\pm$ 1.5 \\
          $\checkmark$ &             $\checkmark$ &             $\checkmark$ &          $\checkmark$ & \textbf{71.8 $\pm$ 0.9} \\
\bottomrule
\end{tabular}

}
\caption{
\textbf{Training Strategies Ablation}.
Average held-out Dice for \method{} models trained with different combinations of the proposed techniques to increase task diversity: in-task augmentation, task augmentation, and synthetic tasks.
}
\label{tab:ablation}
\end{table}

\subpara{Ablation of Training Strategies}.
We perform an ablation study over the three main techniques we employ for increasing data and task diversity during training: in-task augmentation, task augmentation, and synthetic tasks.

Table~\ref{tab:ablation} shows that all proposed strategies lead to improvements in model performance, with the best results achieved when using all strategies jointly, providing a boost of 9 Dice points over no augmentations or synthetic tasks.
Incorporating task augmentation leads to the largest individual improvement of 7.7 Dice points.
Remarkably, the model trained using only synthetic data performs surprisingly well on the medical held-out tasks despite having never been exposed to medical training data.
These results suggest that increasing image and task diversity during training, even artificially, has a substantial effect on how the model generalizes to unseen segmentation tasks.

\begin{figure}[t]
\centerline{\includegraphics[width=\columnwidth]{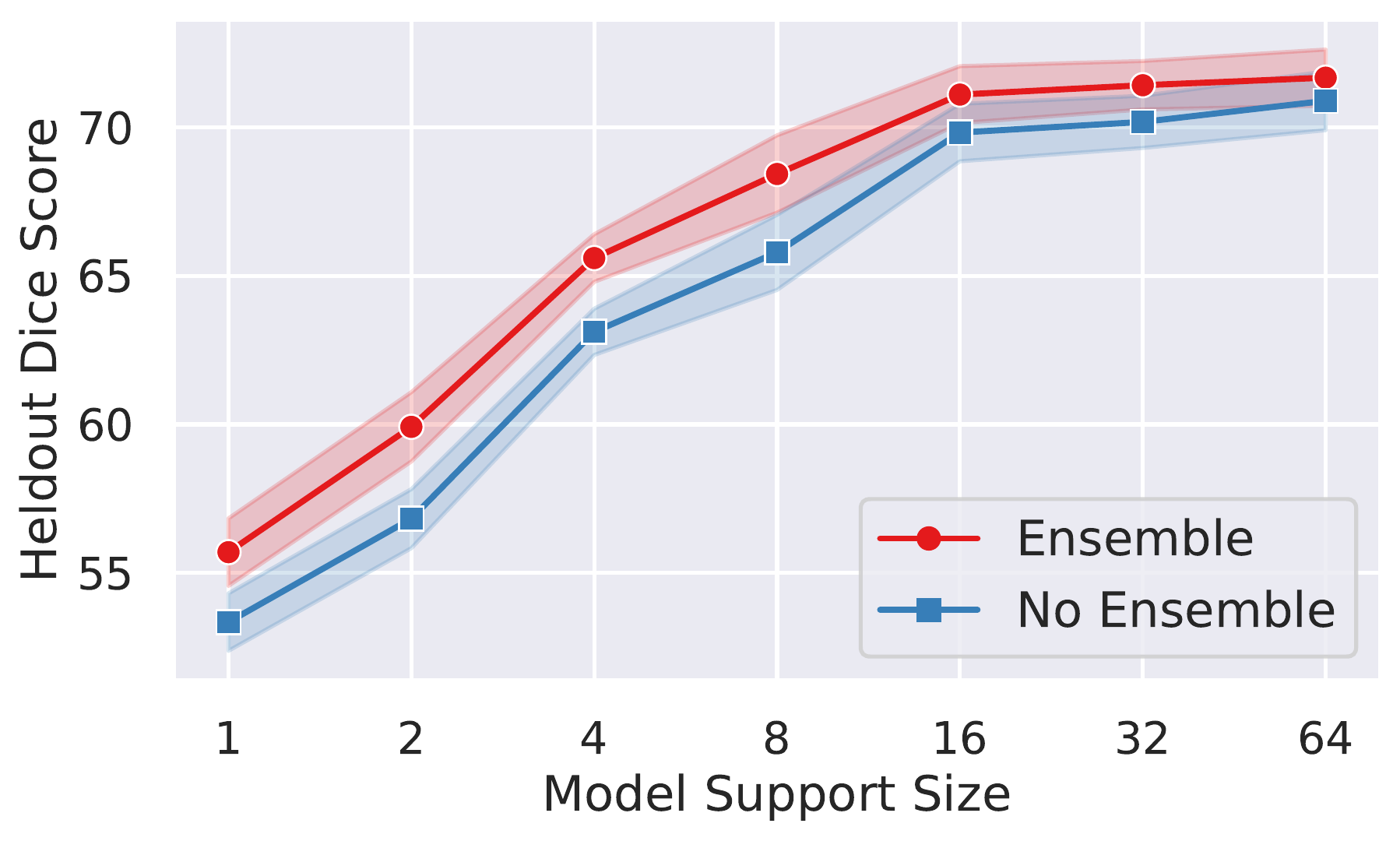}}
\caption{\textbf{Effects of support size.} Relationship between models trained at certain support sizes and their average held-out Dice score. Results improve with higher support size, with ensembling consistently helping.
}
\label{fig: SuppSetSize}
\end{figure}

\subpara{Support Set Size.}
We study the effect of support size on models trained with support sizes~$N$ from 1 to 64.

Figure~\ref{fig: SuppSetSize} shows that the best results are achieved with large training support set sizes, with the average held-out Dice rapidly improving from 53.7 to 69.9 for supports sizes from 1 to 16, and then providing diminishing returns at greater support sizes, with a maximum of 71 Dice at support size 64.
We find that ensembling predictions leads to consistent improvements in all cases, with greater improvements of 2.4-3.1 Dice points
for small support sets ($N < 16$).

\subpara{Limited Example Data}.
Since manually annotating examples from new tasks is expensive for medical data, we investigate how the number of labeled images affects the performance of \method{}.
We study \method{} when using a limited amount of labeled examples~$N$ at inference, for $N = 1,2,\ldots,64$.
We perform 100 repetitions for each size, each corresponding to an independent random subset of the data.
Here, the support set contains all available data for inference, and thus we do not perform ensembling.

\begin{figure}[t]
\centerline{\includegraphics[width=\columnwidth]{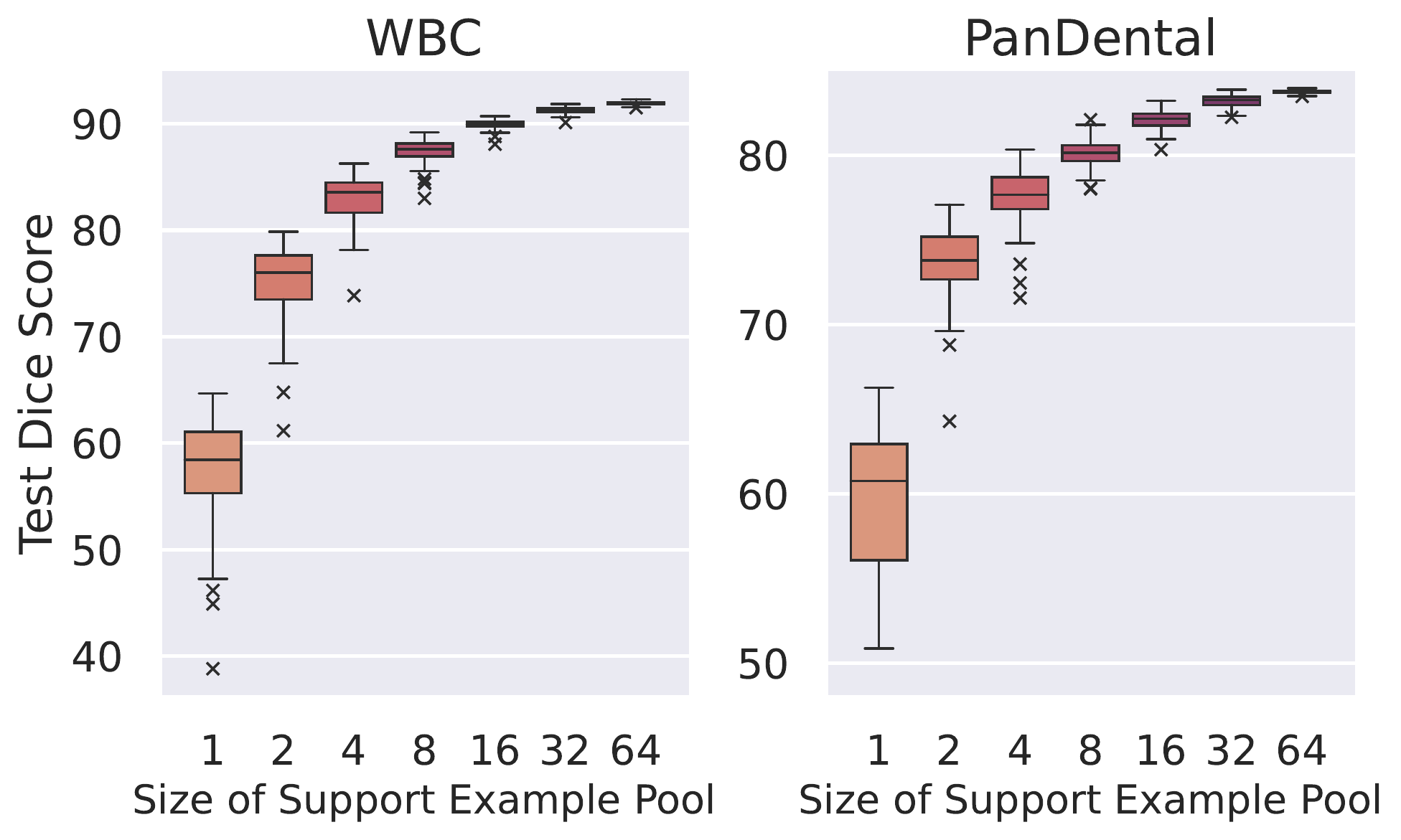}}
\caption{\textbf{Effect of available data at inference.}
\method{} predictions using a limited $d_\text{support}$ example pool on the held-out WBC and PanDental datasets.
For each size, we perform 100 repetitions using different random subsets. 
\vspace{-5pt}
}
\label{fig: DataAtInference}
\end{figure}

Figure~\ref{fig: DataAtInference} presents results for the WBC and PanDental held-out datasets, which have 108 and 116 examples in their $d_\text{support}$ splits respectively.
For small values of support size $N$, we observe a large variance caused by very diverse support sets.
As $N$ increases, we observe that average segmentation quality monotonically improves and the variance from the sample of available data examples is greatly reduced.
We include analogous figures in the supplement for the other held-out datasets, where we find similar trends.

\subpara{Support Set Ensembling.}
We study the effect of varying the support size $N$ at inference, and number $K$ of predictions being ensembled.
We first sample 100 independent support sets for each inference support size $N$.
Then, for each ensembling amount $K$, we compute ensembled predictions by averaging $K$ independently drawn predictions. %

\begin{figure*}
\centering
\includegraphics[width=0.75\textwidth]{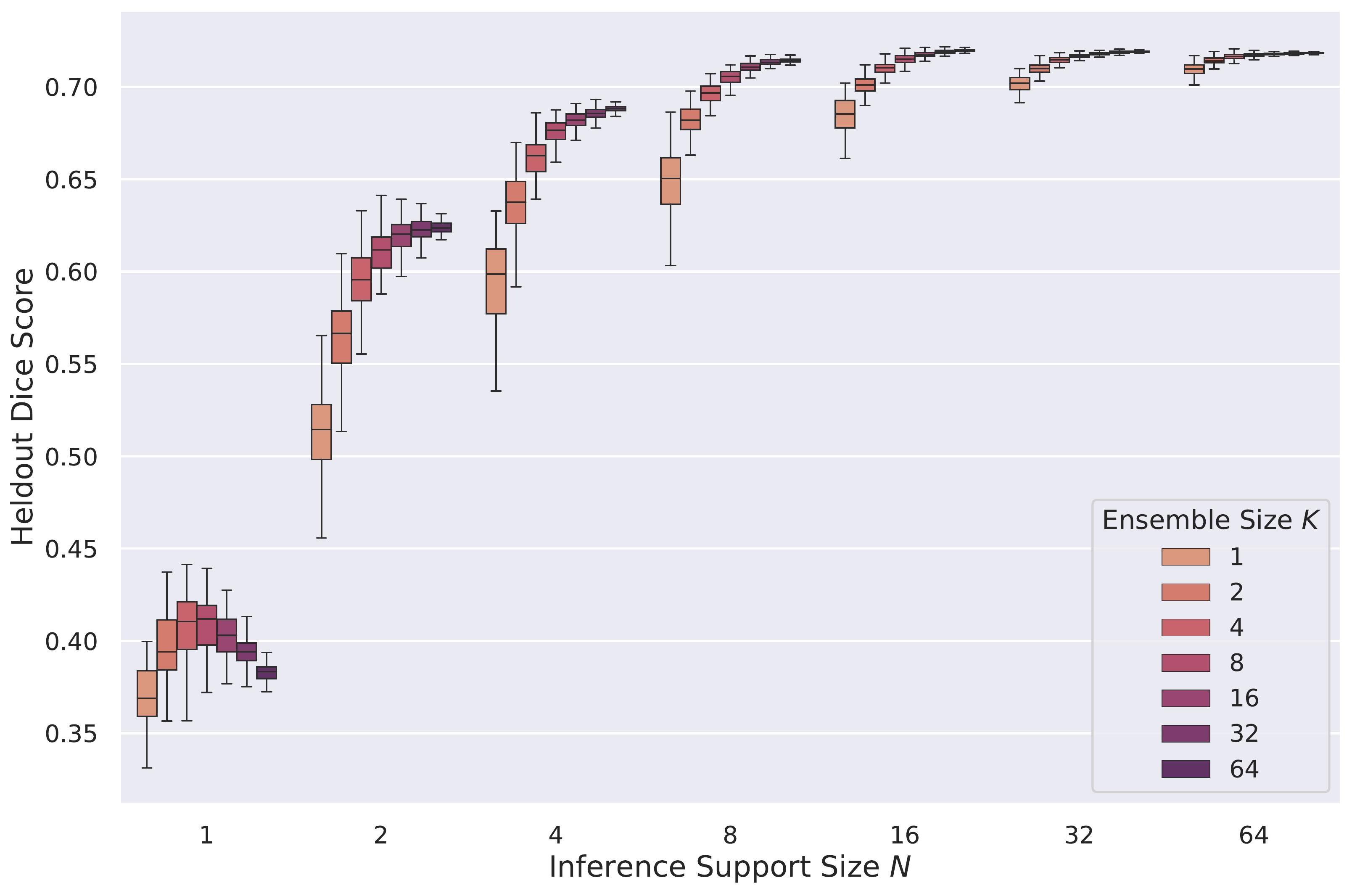}
\caption{%
\textbf{Ensembling predictions at different inference support sizes.}
Average held-out test Dice Score for different settings of ensembling and support size.
For each inference support size $N$, we report the results (in average held-out Dice Score) of taking 100 predictions ($K=1$) and ensembling by averaging in groups of size $K$, performing 100 repetitions for each $K$.
The value boxes report quantiles over the 100 values for each setting and find that increasing either $K$ or $N$ leads to improved model performance, with $N$ having a significantly larger effect than $K$.
}
\label{fig: SupportEnsembling}
\end{figure*}

Figure~\ref{fig: SupportEnsembling} shows that given a certain support size, increasing the ensemble size leads to monotonic improvements and reduced variance, likely by being less dependent on the specific examples in the support set.
The performance also monotonically improves with increased support size~$N$, which has a significantly larger effect on segmentation accuracy than increasing the ensemble size.
For instance, non-ensembled predictions with support size 64 ($N=64$, $K=1$) are better than heavily ensembled predictions with smaller support sizes ($N=2,4,8$ and $K=64$), even though the latter uses more support examples.
This suggests that \method{} models exploit information coming from the support examples in a fundamentally different way than existing ensembling techniques used in FS learning.

\vspace{-2pt}
\section{Discussion and Conclusion}
\vspace{-2pt}

We introduce \method{}, an approach for learning a \emph{single} task-agnostic model for medical image segmentation.
We use a large and diverse collection of open-access medical segmentation datasets to train \method{}, which is capable of generalizing to unseen anatomies and tasks.
We introduce a novel \emph{cross-convolution} operation that interacts the query and support representations at different scales.

In our experiments, \method{} substantially outperforms existing few-shot methods in all held-out datasets.
Through extensive ablation studies, we conclude that \method{} performance is strongly dependent on task diversity during training and support set diversity during inference. This highlights the utility of \method{} facilitating variably-sized support sets, enabling flexibility to potential users' datasets.

\subpara{Limitations.} In this work, we focused on demonstrating and thoroughly analyzing the core idea of \method{}, using 2D data and single labels. We are excited by future extensions to segment 3D volumes using 2.5D or 3D models and multi-label maps, and further closing the gap with the upper bounds.

\subpara{Outlook.} \method{} promises to easily adapt to new segmentation tasks determined by scientists and clinical researchers, without model retraining that is often impractical for them.

{\small
\bibliographystyle{conf_ieee_fullname}
\bibliography{references}
}

\appendix
\clearpage

\onecolumn

\section{MegaMedical}
\label{sec: supp-megamedical}

\subpara{Preprocessing.}  Medical images involve large variations of voxel or pixel values. For example, MRI intensities in MegaMedical range from [0, 800], CT intensities range from [-2000, 2000], while other modalities might already be in the $[0, 1]$ range.

To normalize data across the diverse datasets, we apply several preprocessing steps for each modality.
For MRI datasets, we clip the intensity to~$[0.5, 99.5]$ percentiles for non-zero voxels. For CT images, we clip intensity values to the range $[-500, 1000]$. We min-max normalize all resulting volumes to~$[0,1]$ and resize them to $128\times128\times128$. From any 3D volumes, we extract two different kinds of slices: \textit{mid-slices} and \textit{max-slices}. 

\subpara{Slicing.} For \textit{mid-slices}, from any 3D image and label volumes we extract the middle slice along each axis, resulting in a representative $128 \times 128$ slice. This strategy avoids biasing the data toward knowing the location of labels in the scans. This is especially important for inference, where the location of the foreground label would not be known in a 3D volume.

For training, we also extract \textit{max slices}. For each label $l$ of a dataset, we find the slice (along each axis) in each volume that contains the most voxels with that label. We extract this slice from both volume and label map and repeat this for all labels in the dataset. These slices provide additional training data, and we do not use them during evaluation.

\subpara{Label Maps.} Most datasets include label maps that were either manually obtained, or manually curated after being obtained using an automatic tool. For adult brain datasets~\cite{MCIC, OASIS-data, PPMI}, we follow recent large-scale analyses~\cite{balakrishnan2019voxelmorph, OASIS-proc} and obtain semantic sub-cortical segmentations using FreeSurfer~\cite{fischl2012freesurfer}.

Datasets can often contain multiple tasks -- such as segmenting both lesions and anatomy -- and the same task can appear in different datasets -- like segmenting the hippocampus in different MRI collections. Certain labels can sometimes tackle the same anatomical region of interest but be defined differently in two different datasets. In this work, we focus on single-label, single-modality, and 2D segmentation.

\subpara{Medical Task Creation.} To create a task, the subjects of dataset $d$ can contain labels for either a particular biomedical target (e.g. eye-vessels~\cite{DRIVE}, vertebrae~\cite{SpineWeb}, white blood cells~\cite{WBC}) or a set of targets (e.g. abdominal organs~\cite{CHAOSdata2019}, brain regions~\cite{OASIS-data}), and an imaging modality $m \in M_d$ (e.g. CT, MRI, X-Ray). If $d$ is multi-class, we split it into several single-label $l \in L_d$ tasks. If $d$ is a 3D dataset, we extract different axes $a \in A_d$ as different tasks. Following this construction, each task can be described using a unique tuple $t = (d, m, l, a)$.

\begin{table*}[t]
   \rowcolors{2}{white}{gray!15}
  \caption{We assembled the following set of datasets to train \method{}. For the relative size of datasets, we have included the number of unique scans (subject and modality pairs) that each dataset has.}
  \begin{tabular}{lp{6cm}rl}
    \textbf{Dataset Name }   & \textbf{Description} & \textbf{\# of Scans} & \textbf{Image Modalities} \\
    \toprule
     AbdomenCT-1K~\cite{AbdomenCT-1K} & Abdominal organ segmentation (overlap with KiTS, MSD) & 361 & CT \\
     ACDC~\cite{ACDC} & {Left  and right ventricular endocardium} & 99 & cine-MRI \\
     AMOS~\cite{AMOS} & Abdominal organ segmentation & 240 & CT, MRI \\
     BBBC003~\cite{BBBC003} & Mouse embryos & 15 & Microscopy \\
     BrainDevelopment~\cite{gousias2012magnetic, BrainDevFetal, BrainDevelopment, serag2012construction} & Adult and Neonatal Brain Atlases & 53 & multi-modal MRI \\
     BRATS~\cite{BRATS, bakas2017advancing, menze2014multimodal} & Brain tumors & 6,096 & multi-modal MRI \\
     BTCV~\cite{BTCV} & Abdominal Organs & 30 & CT  \\
     BUS~\cite{Bus} & Breast tumor & 163 & Ultrasound \\
     CAMUS~\cite{CAMUS}  & Four-chamber and Apical two-chamber heart & 500 & Ultrasound\\
     CDemris~\cite{cDemris}  & Human Left Atrial Wall & 60 & CMR \\
     CHAOS~\cite{Chaos_1, Chaos_2}  & Abdominal organs (liver, kidneys, spleen) & 40 & CT, T2-weighted MRI \\
     CheXplanation~\cite{CheXplanation} & Chest X-Ray observations & 170 & X-Ray\\
     CoNSeP~\cite{CoNSeP}  & Histopathology Nuclei & 27 & Microscopy \\
     DRIVE~\cite{DRIVE} & Blood vessels in retinal images & 20 & Optical camera\\
     EOphtha~\cite{EOphtha} & Eye Microaneurysms and Diabetic Retinopathy & 102 & Optical camera\\
     FeTA~\cite{FeTA} & Fetal brain structures & 80 & Fetal MRI \\
     FetoPlac~\cite{FetoPlac} & Placenta vessel & 6 & Fetoscopic optical camera\\
     HMC-QU~\cite{HMC-QU, kiranyaz2020left} & 4-chamber (A4C) and apical 2-chamber (A2C) left  wall & 292 & Ultrasound \\
     I2CVB~\cite{I2CVB} & Prostate (peripheral zone, central gland) & 19 & T2-weighted MRI \\
     IDRID~\cite{IDRID} & Diabetic Retinopathy & 54 & Optical camera \\
     ISLES~\cite{ISLES} & Ischemic stroke lesion & 180 & multi-modal MRI \\
     KiTS~\cite{KiTS} & Kidney and kidney tumor & 210 & CT \\
     LGGFlair~\cite{buda2019association, LGGFlair} & TCIA lower-grade glioma brain tumor & 110 & MRI \\
     LiTS~\cite{LiTS} & Liver Tumor & 131 & CT \\
     LUNA~\cite{LUNA} & Lungs & 888 & CT \\
     MCIC~\cite{MCIC} & Multi-site Brain regions of Schizophrenic patients & 390 & T1-weighted MRI\\
     MSD~\cite{MSD} & Large-scale collection of 10 Medical Segmentation Datasets & 3,225 & CT, multi-modal MRI\\
     NCI-ISBI~\cite{NCI-ISBI} & Prostate & 30 & T2-weighted MRI \\
     OASIS~\cite{OASIS-proc, OASIS-data} & Brain anatomy & 414 & T1-weighted MRI \\
     OCTA500~\cite{OCTA500} & Retinal vascular & 500 & OCT/OCTA \\
     PanDental~\cite{PanDental} & Mandible and Teeth & 215 & X-Ray \\
     PROMISE12~\cite{Promise12} & Prostate & 37 & T2-weighted MRI\\
     PPMI~\cite{PPMI,dalca2018anatomical} & Brain regions of Parkinson patients & 1,130 & T1-weighted MRI\\
     ROSE~\cite{Rose} & Retinal vessel & 117 & OCT/OCTA \\
     SCD~\cite{SCD} & Sunnybrook Cardiac Multi-Dataset Collection & 100 & cine-MRI \\
     SegTHOR~\cite{SegTHOR} & Thoracic organs (heart, trachea, esophagus) & 40 & CT \\
     SpineWeb~\cite{SpineWeb} & Vertebrae & 15 & T2-weighted MRI  \\
     STARE~\cite{STARE} & Blood vessels in retinal images & 20 & Optical camera \\
     TUCC~\cite{TUCC} & Thyroid nodules & 167 & Ultrasound cine-clip \\
     WBC~\cite{WBC} & White blood cell and nucleus & 400 & Microscopy \\
     WMH~\cite{WMH} & White matter hyper-intensities & 60 & multi-modal MRI \\
     WORD~\cite{Word} & Organ segmentation & 120 & CT \\
     \bottomrule
  \end{tabular}
  \label{tab: datasets}
\end{table*}

\clearpage
\section{Additional Implementation Details}
\label{sec:impl}

\textbf{Data Storage}.
For each gradient step, a UniverSeg model needs to load $B\times (N+1) \times 2$ images, where $B$ is the batch size, $N$ is the support size, and the factor of 2 corresponds to the combination of the image and label map.
This can pose a serious challenge for traditional data loading strategies, especially as $N$ increases.
Therefore, we store data samples in a highly optimized way to ensure that I/O does not bottleneck the training process, using LMDB data stores that are optimized for read-only access.
Within the database, data is encoded using msgpack and compressed with the LZ4 codec for fast decompression.
We find that this setup exceeds regular file-system random access by over two orders of magnitude.

\textbf{Task Sampling}. To ensure task and data heterogeneity during training, we do not sample all tasks equally. Some datasets contain substantially more tasks than others, and we aim to avoid overfitting medical domains where tasks are abundant (such as neuroimaging tasks).
Instead, we perform hierarchical uniform sampling with multiple stages: dataset, subject group, acquisition modality, axis, and label. We first sample the dataset uniformly from all datasets, then sample a task among the tasks from that dataset, and so on.

\textbf{Model}.
We implemented \method{} in PyTorch~\cite{paszke2019pytorch} and used the official implementations for the baselines (ALPNet, PANet, and SENet) and supervised network nnUNet. Based on the experimental details in the ALPNet work, we used an off-the-shelf ResNet101~\cite{he2016deep} for both the pre-trained encoder for ALPNet and PANet. For these two methods, because their feature encoder expects three-channel inputs, we duplicate the input dimension $1 \times 128 \times 128$ three times channel-wise to get inputs of dimension $3 \times 128 \times 128$.

We efficiently perform the CrossConvolution operation by exploiting the batch dimension. Instead of performing $N$ convolutions with the same learnable parameters, we perform a single convolution by tiling the inputs along the batch dimension.
We use the same strategy for the convolutions predicting the CrossBlock outputs~$V^\prime$.

\textbf{Optimization}.
For all models during training, we minimize the soft Dice loss:
 \begin{equation}
    \begin{aligned}
      \mathcal{L}_\text{Dice}(y_t, \hat{y})  &= \frac{2\sum y_t \odot \hat{y}}{\sum y_t^2 +  \sum \hat{y}^2},
    \end{aligned}
    \label{eq:soft-Dice}
\end{equation}
using a learning rate of $\eta = 10^{-4}$, the Adam optimizer\cite{kingma2014adam}, and a batch size of 1.
We searched learning rates over the range $[10^{-5}, 10^{-2}]$ and found the best results on the validation split of the training datasets with learning rates around $10^{-4}$ and set on $10^{-4}$ for comparison and reproducibility purposes.

\textbf{Evaluation}.
We evaluate predicted label maps~$\hat{y}$ using the Dice score~\cite{dice1945measures}, which quantifies the overlap between two regions and is widely used in the segmentation literature:
\begin{equation}
    \begin{aligned}
      \text{Dice}(y_t, \hat{y})  &= 100 * \frac{2|y_t \cap \hat{y}|}{|y_t|^2 + |\hat{y}|^2}
    \end{aligned}
    \label{eq:Dice}
\end{equation}
where~$y$ is the ground truth segmentation map and~$\hat{y}$ is the predicted segmentation map.
A Dice score of 100 indicates perfectly overlapping regions, while 0 indicates no overlap.

\textbf{Task-Specific Networks} The nnUNet framework trains 5 networks per task using multiple folds of the support data for training, and ensemble their predictions at inference. We apply the nnUNet framework independently for each held-out task, which corresponds to a set of subjects and the segmentation labels for a particular binary task.

We also designed and trained additional individual U-Net networks. For the majority of the tasks, we found the best results after searching batch sizes and augmentation policies. We omitted these as we found that the nnUNets performed very similarly.

\clearpage
\section{Data Augmentation}
\label{sec: supp-aug}

During \method{} training, we found that using substantial data augmentation was important. Augmentation techniques enable \method{} to see effectively both a greater diversity of tasks as well as a greater number of examples of each. We separate these two kinds of augmentations into \textit{Task} and \textit{In-Task}.

In Table~\ref{tab: augmentations}, we detail included augmentations. During model development, we experimented to find the hyperparameters which worked best for each kind of augmentation. Several augmentations are repeated (although with different parameters) across task and in-task sections of Table~\ref{tab: augmentations}. %

\begin{table}[H]
   \rowcolors{2}{gray!15}{white}
  \caption{\textbf{List of augmentations used in model training.}}
  \begin{tabular}{lll}
    \textbf{Augmentation} &\textbf{ Aug Type} & \textbf{Parameter Details}\\
    \toprule
     Flip Intensities                   & Task       & $\textbf{p} = 0.50$ \\
     Flip Labels                        & Task       & $\textbf{p} = 0.50$ \\
     Horizontal/Vertical Flip           & Task       & $\textbf{p} = 0.50$ \\
     Sobel-Edge Label                   & Task       & $\textbf{p} = 0.50$ \\
     Task Affine Shift                  & Task       & $\textbf{p} = 0.50$ $\textbf{\text{degrees}} = [0, 360]$ $\textbf{\text{translate}} = [0, 0.2]$ $\textbf{\text{scale}} = [0.8, 1.1]$ \\
     Task Brightness Contrast Change    & Task       & $\textbf{p} = 0.50$ $\textbf{\text{brightness}} = [-0.1, 0.1]$ $\textbf{\text{contrast}} = [0.8, 1.2]$  \\
     Task Elastic Warp                  & Task       & $\textbf{p} = 0.25$ $\boldsymbol{\alpha} = [1, 2]$ $\boldsymbol{\sigma} = [6, 8]$ \\
     Task Gaussian Blur                 & Task       & $\textbf{p} = 0.50$ $\textbf{\text{k-size}} = 5$ $\boldsymbol{\sigma} = [0.1, 1.1]$ \\
     Task Gaussian Noise                & Task       & $\textbf{p} = 0.50$ $\boldsymbol{\mu} = [0, 0.05]$ $\boldsymbol{\sigma^2} = [0, 0.05]$ \\
     Task Sharpness Change              & Task       & $\textbf{p} = 0.50$ $\textbf{\text{sharpness}} = 5$ \\
     \hline
     Example Affine Shift               & In-Task  & $\textbf{p} = 0.50$ $\textbf{\text{degrees}} = [0, 360]$ $\textbf{\text{translate}} = [0, 0.2]$ $\textbf{\text{scale}} = [0.8, 1.1]$ \\
     Example Brightness Contrast Change & In-Task  & $\textbf{p} = 0.25$ $\textbf{\text{brightness}} = [-0.1, 0.1]$ $\textbf{\text{contrast}} = [0.5, 1.5]$  \\
     Example Gaussian Blur              & In-Task  & $\textbf{p} = 0.25$ $\textbf{\text{k-size}} = 5$ $\boldsymbol{\sigma} = [0.1, 1.1]$ \\
     Example Gaussian Noise             & In-Task  & $\textbf{p} = 0.25$ $\boldsymbol{\mu} = [0, 0.05]$ $\boldsymbol{\sigma^2} = [0, 0.05]$ \\
     Example Sharpness Change           & In-Task  & $\textbf{p} = 0.25$ $\textbf{\text{sharpness}} = 5$ \\
     Example Variable Elastic Warp      & In-Task  & $\textbf{p} = 0.80$ $\boldsymbol{\alpha} = [1, 2.5]$ $\boldsymbol{\sigma} = [7, 8]$ \\
     \bottomrule
  \end{tabular}
  \label{tab: augmentations}
\end{table}

\begin{figure*}
\centerline{\includegraphics[width=0.8\columnwidth]{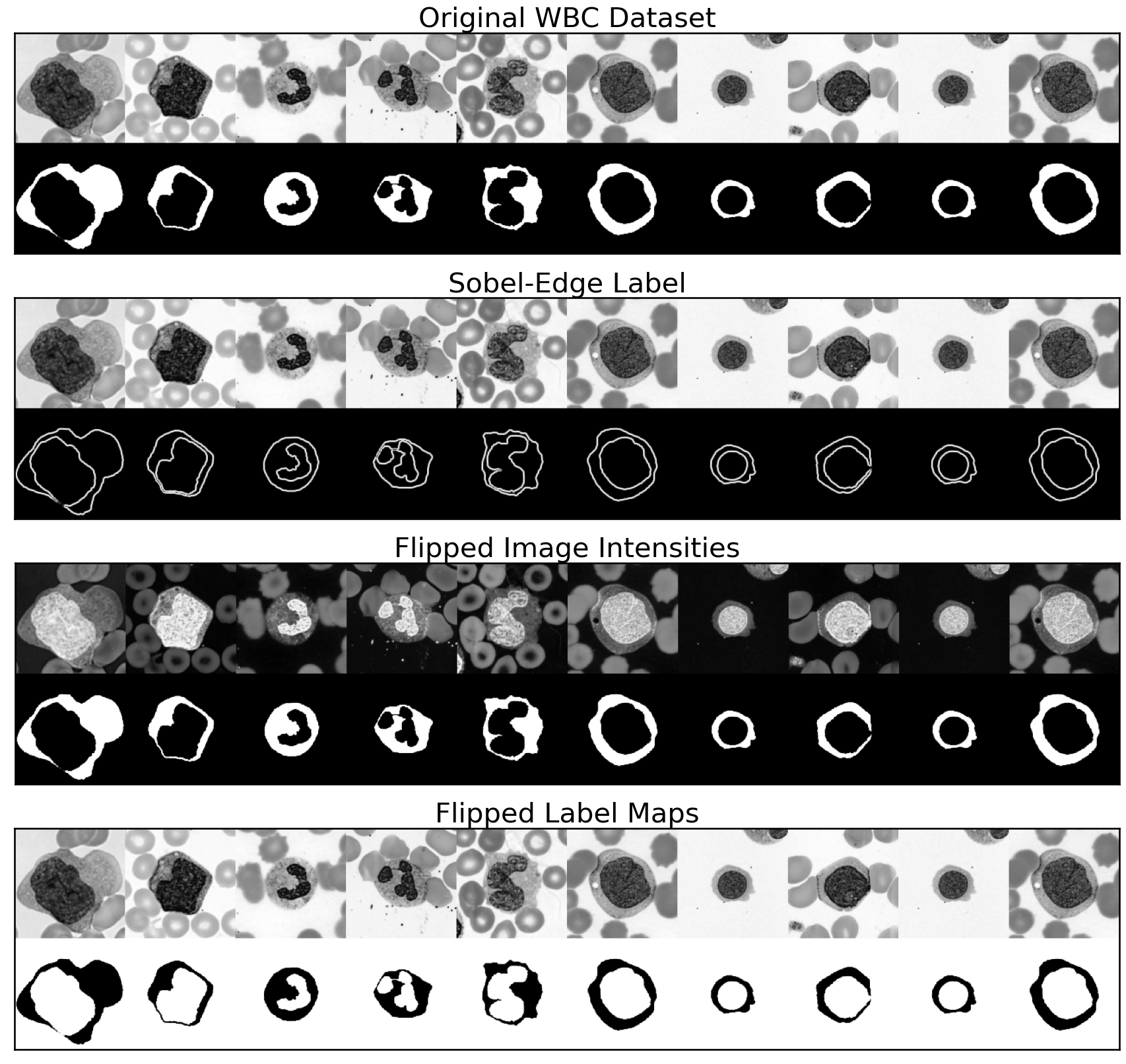}}
\caption{\textbf{Example augmentation operations applied to the WBC Dataset.} We visualize several examples of unique task augmentations we apply during training.}
\label{supp: aug-fig}
\end{figure*}

We briefly describe each augmentation and its parameters. Each augmentation also has a parameter $\textbf{p}$ which controls the probability that augmentation is applied at each iteration. For in-task augmentation, this probability controls whether or not \textbf{all} of the support set entries are individually augmented or not. For operations that we developed, we include examples in Figure~\ref{supp: aug-fig}.

\begin{itemize}
    \item \textit{Flip Intensities} (Task): Flip the intensity values for all images (query and support), but not the label maps, using 1 - image for each.
    \item \textit{Flip Labels} (Task): Reverse the foreground and background in the segmentation maps. 
    \item \textit{Horizontal/Vertical Flip} (Task): Flip all entries in the support horizontally or vertically (all flipped in the same way).
    \item \textit{Sobel-Edge Label} (Task): We propose an operation that increases the number of tasks with thin segmentation structures. We apply a Sobel filter to each label map in the x and y directions, compute the squared norm, which becomes our new label map.
    \item \textit{Affine Shift} (Task, In-Task): Apply a consistent random affine transformation to all entries in the support set; \textbf{degrees} controls how much to randomly rotate, \textbf{translate} controls how far the images and labels can shift, and \textbf{scale} controls the amount of zoom.
    \item \textit{Brightness Contrast Change} (Task, In-Task): Apply a random brightness and contrast change to all images; how much brightness can change is controlled by \textbf{brightness} and contrast is controlled by the parameter \textbf{contrast}.
    \item \textit{Elastic Warp} (Task, In-Task): Apply a consistent elastic deformation warp to all entries in the support and to the query; $\boldsymbol{\alpha}$ controls the strength of the warp and $\boldsymbol{\sigma}$ controls the smoothness of the warp.
    \item \textit{Gaussian Blur} (Task, In-Task): Apply a convolutional Gaussian blur to each image in the support set and the query with a certain kernel size, \textbf{k-size}, and standard-deviation $\boldsymbol{\sigma}$.
    \item \textit{Gaussian Noise} (Task, In-Task): Apply Gaussian noise to all images in the support set and query with mean $\boldsymbol{\mu}$ and variance $\boldsymbol{\sigma^2}$.
    \item \textit{Sharpness Change} (Task, In-Task): Apply a sharpness filter to the images (query and support), where the sharpness strength is controlled by \textbf{sharpness}.
\end{itemize}

\clearpage
\section{Synthetic Tasks}
\label{sec: supp-synth}

\setcounter{figure}{10}

\begin{figure*}[!h]
\centerline{\includegraphics[width=0.8\columnwidth]{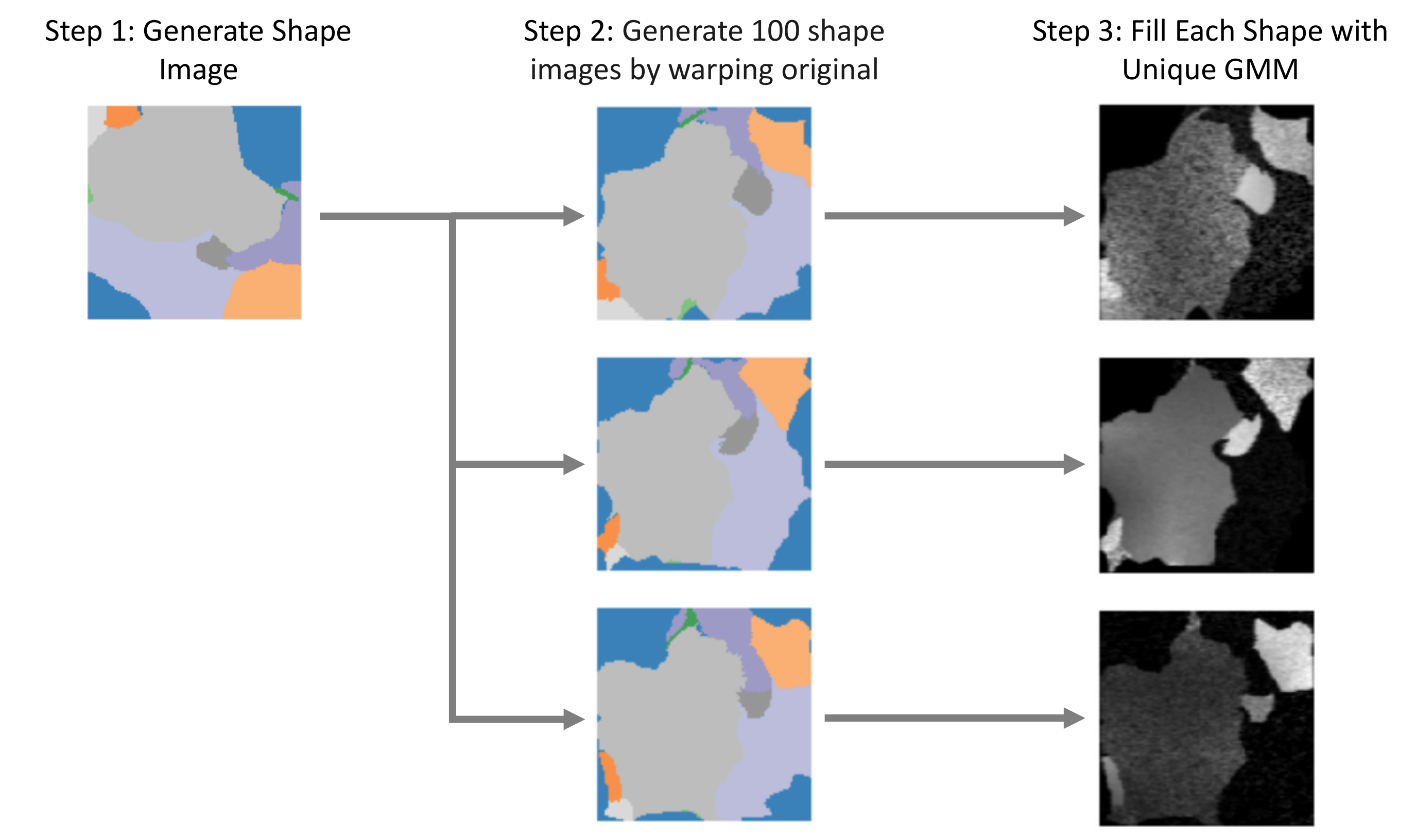}}
\caption{\textbf{Generation process for synthetic tasks.} For a new synthetic task, we first generate random shapes to obtain a label map, then synthesize 100 spatial variations on this label map, and finally synthesize resulting intensity images. We repeat this process for 1000 tasks.}
\label{fig: SyntheticTasks-gen}
\end{figure*}

\begin{figure*}[!h]
\centerline{\includegraphics[width=\columnwidth]{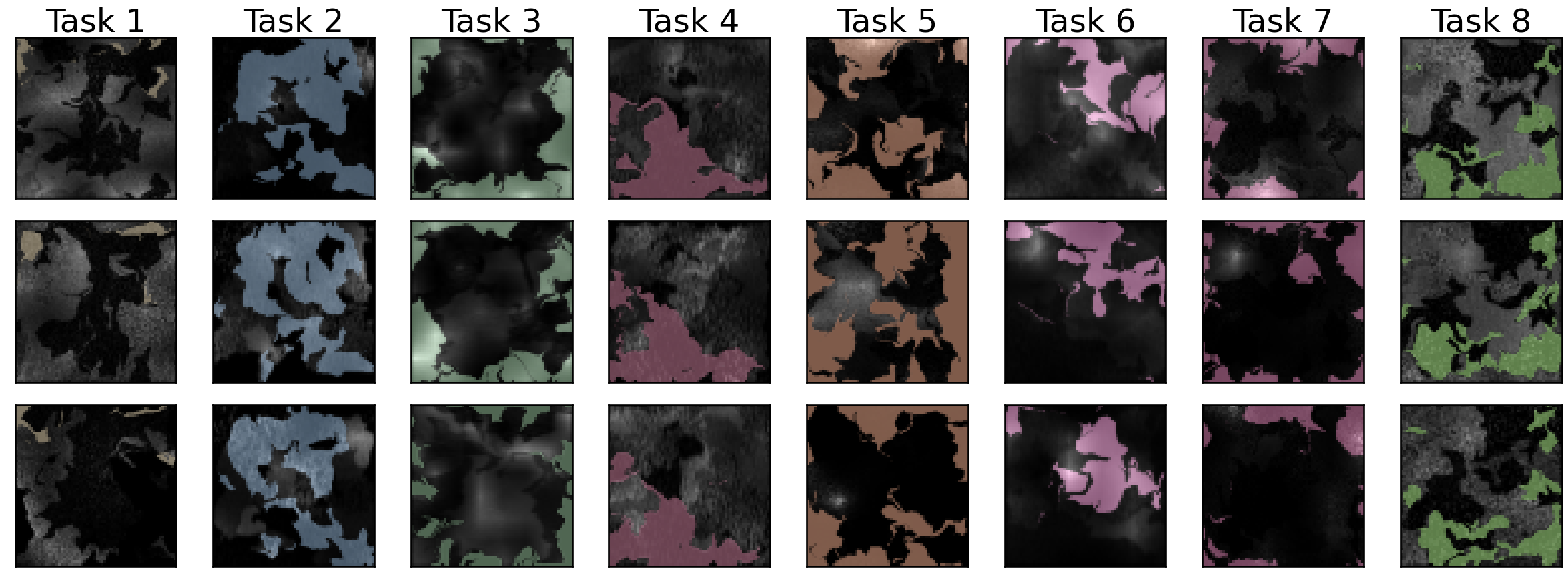}}
\caption{\textbf{Examples of Synthetically Generated Tasks.} We visualize 10 of the 1000 synthetically generated tasks, involving varying shapes, textures, and label shapes.}
\label{fig: SyntheticTasks}
\end{figure*}

We found improvement in held-out performance by introducing synthetic tasks during training, building on recent methods that use synthetic medical images to solve specific tasks~\cite{billot2020learning,hoffmann2022synthmorph,hoopes2022synthstrip}, especially the synthetic shapes in SynthMorph~\cite{hoffmann2022synthmorph}. We generate 1,000 new tasks with high diversity (Figure~\ref{fig: SyntheticTasks}). As shown in Figure~\ref{fig: SyntheticTasks-gen}, for each task, we first synthesize a label map of 16 random shapes, representing 16 regions of interest. We deform this label map with 100 random smooth deformation fields, representing 100 subjects with the same simulated anatomy. We then add texture to the resulting images by filling in each region of interest with slightly varied intensities around a sampled mean and adding Gaussian and Perlin noise.

\clearpage

\section{Extended Results}
\label{sec: supp-ext-results}

\subsection{Main Results}
We include detailed numbers corresponding to figures in the main body of the paper.
\begin{itemize}
\item \textbf{Method Comparison}. Table~\ref{tab:main-results-per-dataset} reports test performance numbers of the results from Figure~\ref{tab:main-results} and Table~\ref{tab:main-results}, comparing the segmentation results of \method{} to the FS baselines and the supervised nnUNet upper bounds.
\item \textbf{Training Strategies Ablation}. Table~\ref{tab:aug-ablation-per-dataset} reports per-dataset test performance numbers for the results of Table~\ref{tab:ablation} comparing several ways of augmenting the task diversity artificially. While the overall trend holds for most datasets, we find that the increase in task diversity has a detrimental effect on the STARE eye vessel segmentation task.
\item \textbf{Model Support Size}. Table~\ref{tab:support-size-per-dataset} reports held-out test performance numbers of the results from Figure~\ref{fig: SuppSetSize} along with per-dataset breakdowns. We find that the global trend holds for each individual dataset, with larger support sizes achieving better results and ensembling (with $K=10$) consistently improving predictions.
\item \textbf{Available Data for Inference Ablation}. Table~\ref{tab:inference-size-per-task} reports extended results from Figure~\ref{fig: DataAtInference} with per-task results as we change the size of the support example pool. All tasks showcase the same trend with consistent improvements as more support examples are used during inference and with a reduced variance across random subsets of the support split.
\item \textbf{Support Set Ensembling}.
 Table~\ref{tab:ensemble-results} reports results for the support set ensembling experiment. We observe a clear difference between $N=1$ and $N>1$ for ensembled predictions. For $N=1$ ensembling leads to small improvements that eventually decline as $K$ grows. In contrast for $N>1$, ensembling leads to substantial improvements that also reduce the variance of the distribution, limiting the dependence on the specific subset used for the support set.
\item \textbf{Number of Tasks Ablation}. Table~\ref{tab:numtasks} reports the per-dataset and global dice numbers for the models trained with a subset of the training datasets.
\end{itemize}

\begin{table}[!h]
\centering
\caption{%
\textbf{Method Comparison}.
Test Dice Score for the baselines, UniverSeg, and the nnUNet upper bounds in each of the held-out datasets.
Standard deviation is computed by bootstrapping subjects before hierarchically averaging the data.
}
\begin{tabular}{lcccccc@{\hskip 0.4in}c}
     \textbf{Model} &       \textbf{ACDC} &  \textbf{PanDental} &        \textbf{SCD} &      \textbf{STARE} &   \textbf{SpineWeb} &        \textbf{WBC} &       \textbf{All (avg.)} \\
\toprule
          ALPNet &       34.6 $\pm$ 2.4 &       72.9 $\pm$ 0.8 &       53.4 $\pm$ 3.0 &       17.8 $\pm$ 1.9 &       31.6 $\pm$ 4.6 &       76.2 $\pm$ 1.1 &       47.8 $\pm$ 1.1 \\
           PANet &       27.8 $\pm$ 4.3 &       67.7 $\pm$ 0.8 &       58.9 $\pm$ 3.4 &       20.1 $\pm$ 3.2 &       21.8 $\pm$ 0.4 &       54.7 $\pm$ 1.6 &       41.8 $\pm$ 1.3 \\
           SENet &       40.1 $\pm$ 2.0 &       81.1 $\pm$ 0.9 &       55.4 $\pm$ 3.3 &       35.2 $\pm$ 2.2 &       18.3 $\pm$ 4.0 &       70.8 $\pm$ 1.3 &       50.1 $\pm$ 1.3 \\
UniverSeg (ours) & \textbf{70.9 $\pm$ 2.9} & \textbf{87.5 $\pm$ 0.9} & \textbf{69.0 $\pm$ 2.9} & \textbf{48.1 $\pm$ 2.0} & \textbf{64.6 $\pm$ 5.4} & \textbf{90.6 $\pm$ 1.1} & \textbf{71.8 $\pm$ 0.9} \\
   \midrule
   nnUNet (sup.) &       82.5 $\pm$ 2.3 &       92.9 $\pm$ 1.1 &       75.0 $\pm$ 3.4 &       65.5 $\pm$ 1.1 &       91.2 $\pm$ 2.3 &       95.1 $\pm$ 0.7 &       84.4 $\pm$ 1.0 \\
\bottomrule
\end{tabular}

\label{tab:main-results-per-dataset}
\end{table}

\begin{table}[!h]
\centering
\rowcolors{2}{gray!15}{white}
\caption{
\textbf{Training Stategies Ablation}.
Per dataset held-out Dice for \method{} models trained with different combinations of the proposed techniques to increase task diversity: in-task augmentation, task augmentation, and synthetic tasks.
}
{\renewcommand{\arraystretch}{1.2}
\begin{tabular}{cccccccccc}
\textbf{Synth} & \textbf{Medical} & \textbf{In-Task} & \textbf{Task} &       \textbf{ACDC} &  \textbf{PanDental} &        \textbf{SCD} &      \textbf{STARE} &   \textbf{SpineWeb} &        \textbf{WBC} \\
\toprule
          $\checkmark$ &               &               &            &       55.4 $\pm$ 3.4 &       80.6 $\pm$ 1.3 &       55.7 $\pm$ 2.4 &       42.6 $\pm$ 2.5 &       50.1 $\pm$ 6.5 &       86.0 $\pm$ 1.4 \\
            &             $\checkmark$ &               &            &       44.9 $\pm$ 1.8 &       85.3 $\pm$ 0.9 &       59.9 $\pm$ 1.9 & \textbf{63.8 $\pm$ 0.9} &       40.3 $\pm$ 6.0 &       82.0 $\pm$ 1.6 \\
          $\checkmark$ &             $\checkmark$ &               &            &       50.6 $\pm$ 2.9 &       85.7 $\pm$ 0.9 &       59.0 $\pm$ 1.9 &       61.9 $\pm$ 1.6 &       45.6 $\pm$ 4.8 &       84.2 $\pm$ 1.4 \\
            &             $\checkmark$ &             $\checkmark$ &            &       52.3 $\pm$ 4.3 &       86.5 $\pm$ 0.9 &       64.9 $\pm$ 2.7 &       56.0 $\pm$ 2.3 &       57.2 $\pm$ 3.7 &       85.1 $\pm$ 1.4 \\
            &             $\checkmark$ &               &          $\checkmark$ &       68.0 $\pm$ 3.0 & \textbf{87.5 $\pm$ 1.0} &       63.5 $\pm$ 2.3 &       56.6 $\pm$ 2.1 &       57.8 $\pm$ 6.6 &       89.2 $\pm$ 1.3 \\
            &             $\checkmark$ &             $\checkmark$ &          $\checkmark$ & \textbf{70.0 $\pm$ 2.8} & \textbf{88.0 $\pm$ 0.9} & \textbf{71.2 $\pm$ 3.1} &       42.2 $\pm$ 2.1 &       58.4 $\pm$ 8.5 & \textbf{90.3 $\pm$ 1.2} \\
          $\checkmark$ &             $\checkmark$ &             $\checkmark$ &          $\checkmark$ & \textbf{70.9 $\pm$ 2.9} & \textbf{87.5 $\pm$ 0.9} & \textbf{69.0 $\pm$ 2.9} &       48.1 $\pm$ 2.0 & \textbf{64.6 $\pm$ 5.4} & \textbf{90.6 $\pm$ 1.1} \\
\bottomrule
\end{tabular}

}
\label{tab:aug-ablation-per-dataset}
\end{table}

\begin{table}
\centering
\caption{
\textbf{Model Support Size}.
Comparison of predictions for models trained with various of support sizes $N$ and evaluated with and without ensembling $K=10$ predictions.
We report results on each held-out dataset as well as the global average.
Standard deviation is computed by bootstrapping subjects before hierarchically averaging the data.
For all datasets, we find that increasing the support size leads to better predictions, with diminishing returns after $N>16$.
Ensembling predictions significantly improve performance in the majority of settings (paired t-test).
}
{\renewcommand{\arraystretch}{1.2}
\begin{tabular}{cccccccc@{\hskip 0.4in}c}
 \textbf{N} & \textbf{K} &        \textbf{ACDC} &   \textbf{PanDental} &         \textbf{SCD} &       \textbf{STARE} &    \textbf{SpineWeb} &         \textbf{WBC} &        \textbf{All (avg.)} \\
\toprule
\multirow{2}{*}{1} & 1  &        41.3 $\pm$ 1.3 &        76.3 $\pm$ 0.9 &        60.2 $\pm$ 1.8 &        37.4 $\pm$ 3.8 &        30.4 $\pm$ 5.5 &        74.0 $\pm$ 1.2 &        53.3 $\pm$ 1.0 \\
   & 10 &        44.5 $\pm$ 2.4 &        79.1 $\pm$ 1.0 &        60.0 $\pm$ 1.9 &        38.5 $\pm$ 4.0 &        32.4 $\pm$ 6.6 &        79.4 $\pm$ 1.4 &        55.7 $\pm$ 1.1 \\
\cline{1-9}
\multirow{2}{*}{2} & 1  &        41.3 $\pm$ 2.6 &        80.0 $\pm$ 1.0 &        63.5 $\pm$ 2.0 &        40.4 $\pm$ 2.1 &        38.0 $\pm$ 4.3 &        77.6 $\pm$ 1.1 &        56.8 $\pm$ 1.0 \\
   & 10 &        42.8 $\pm$ 3.2 &        82.4 $\pm$ 1.1 &        68.0 $\pm$ 2.5 &        40.7 $\pm$ 2.3 &        43.4 $\pm$ 4.1 &        82.3 $\pm$ 1.4 &        60.0 $\pm$ 1.2 \\
\cline{1-9}
\multirow{2}{*}{4} & 1  &        53.9 $\pm$ 1.9 &        83.9 $\pm$ 1.0 &        64.7 $\pm$ 1.7 &        47.9 $\pm$ 2.9 &        45.5 $\pm$ 4.0 &        82.7 $\pm$ 1.4 &        63.1 $\pm$ 0.8 \\
   & 10 &        57.0 $\pm$ 2.6 &        84.6 $\pm$ 1.1 &        66.4 $\pm$ 2.8 &        48.6 $\pm$ 2.9 &        50.8 $\pm$ 4.1 &        85.7 $\pm$ 1.5 &        65.5 $\pm$ 0.8 \\
\cline{1-9}
\multirow{2}{*}{8} & 1  &        57.0 $\pm$ 2.5 &        85.0 $\pm$ 0.9 &        66.9 $\pm$ 3.2 &        45.9 $\pm$ 3.5 &        57.3 $\pm$ 6.5 &        83.7 $\pm$ 1.5 &        66.0 $\pm$ 1.3 \\
   & 10 &        61.6 $\pm$ 3.3 &        86.1 $\pm$ 0.9 &        69.0 $\pm$ 4.1 &        47.1 $\pm$ 3.5 &        62.3 $\pm$ 6.0 &        85.9 $\pm$ 1.5 &        68.6 $\pm$ 1.3 \\
\cline{1-9}
\multirow{2}{*}{16} & 1  &        64.1 $\pm$ 2.4 &        86.1 $\pm$ 0.9 &        69.1 $\pm$ 3.1 &        48.8 $\pm$ 3.0 &        64.4 $\pm$ 5.8 &        86.9 $\pm$ 1.4 &        69.9 $\pm$ 1.0 \\
   & 10 &        66.8 $\pm$ 2.5 &        86.7 $\pm$ 0.9 &        68.7 $\pm$ 3.5 &  \textbf{49.7 $\pm$ 2.8} &  \textbf{66.8 $\pm$ 5.7} &        88.3 $\pm$ 1.5 &        71.2 $\pm$ 1.0 \\
\cline{1-9}
\multirow{2}{*}{32} & 1  &        65.6 $\pm$ 3.0 &        87.1 $\pm$ 0.9 &        69.0 $\pm$ 2.0 &        45.7 $\pm$ 2.2 &        65.8 $\pm$ 4.6 &        87.6 $\pm$ 1.3 &        70.1 $\pm$ 0.9 \\
   & 10 &  \textbf{69.3 $\pm$ 2.9} &  \textbf{87.6 $\pm$ 0.9} &  \textbf{69.5 $\pm$ 1.9} &  \textbf{46.4 $\pm$ 2.1} &  \textbf{66.4 $\pm$ 4.3} &        88.9 $\pm$ 1.4 &  \textbf{71.4 $\pm$ 0.8} \\
\cline{1-9}
\multirow{2}{*}{64} & 1  &        69.0 $\pm$ 2.9 &        87.2 $\pm$ 0.9 &        68.7 $\pm$ 2.9 &        47.2 $\pm$ 2.2 &        64.2 $\pm$ 5.5 &        89.7 $\pm$ 1.1 &        71.0 $\pm$ 1.0 \\
   & 10 &  \textbf{70.9 $\pm$ 2.9} &  \textbf{87.5 $\pm$ 0.9} &  \textbf{69.0 $\pm$ 2.9} &  \textbf{48.1 $\pm$ 2.0} &  \textbf{64.6 $\pm$ 5.4} &  \textbf{90.6 $\pm$ 1.1} &  \textbf{71.8 $\pm$ 0.9} \\
\bottomrule
\end{tabular}

}
\label{tab:support-size-per-dataset}
\end{table}

\begin{table}
\centering
\rowcolors{2}{gray!15}{white}
\caption{
\textbf{Limited Example Data}.
\method{} predictions using a limited $d_\text{support}$ example pool for each held-out task.
For each size, we perform 100 repetitions using different random subsets, reporting the mean and standard deviation across them.
Since some tasks do not have enough subjects to be evaluated for all values of $N$, we report $\min(N, |d_\text{support}|)$ and omit repeated settings where $N > |d_\text{support}|$.
}
{\renewcommand{\arraystretch}{1.2}
\begin{tabular}{lrrrrrrr}
      \textbf{Task} & \textbf{N = 1} & \textbf{N = 2} & \textbf{N = 4} & \textbf{N = 8} & \textbf{N = 16} & \textbf{N = 32} & \textbf{N = 64} \\
\toprule
            ACDC &  22.9 $\pm$ 5.5 &  38.5 $\pm$ 6.9 &  51.4 $\pm$ 4.7 &  59.1 $\pm$ 3.0 &   64.4 $\pm$ 2.2 &   68.6 $\pm$ 1.4 &   71.0 $\pm$ 0.0 \\
PanDental\textsubscript{0} &  59.1 $\pm$ 7.4 &  73.3 $\pm$ 4.2 &  77.6 $\pm$ 1.6 &  80.1 $\pm$ 0.8 &   82.1 $\pm$ 0.5 &   83.2 $\pm$ 0.3 &   83.7 $\pm$ 0.1 \\
PanDental\textsubscript{1} &  65.5 $\pm$ 3.9 &  84.1 $\pm$ 2.2 &  87.5 $\pm$ 2.7 &  89.5 $\pm$ 1.2 &   90.6 $\pm$ 0.5 &   91.1 $\pm$ 0.3 &   91.3 $\pm$ 0.0 \\
      SCD\textsubscript{0} &  34.3 $\pm$ 9.2 &  63.1 $\pm$ 5.1 &  70.8 $\pm$ 2.5 &  73.1 $\pm$ 1.2 &   74.3 $\pm$ 0.4 &   74.2 $\pm$ 0.0 &              \\
      SCD\textsubscript{1} & 33.0 $\pm$ 10.4 &  61.8 $\pm$ 9.4 &  72.8 $\pm$ 5.0 &  76.7 $\pm$ 2.4 &   78.5 $\pm$ 0.8 &   78.6 $\pm$ 0.0 &              \\
      SCD\textsubscript{2} & 45.0 $\pm$ 11.4 & 71.5 $\pm$ 12.7 &  80.6 $\pm$ 7.2 &  84.8 $\pm$ 0.0 &              &              &              \\
      SCD\textsubscript{3} &  30.5 $\pm$ 9.3 &  47.1 $\pm$ 6.5 &  54.9 $\pm$ 4.5 &  63.0 $\pm$ 2.3 &   64.2 $\pm$ 0.0 &              &              \\
      SCD\textsubscript{4} &   9.2 $\pm$ 4.4 &  13.3 $\pm$ 8.3 &  25.9 $\pm$ 7.8 &  39.0 $\pm$ 3.1 &   41.0 $\pm$ 0.0 &              &              \\
           STARE &  25.5 $\pm$ 3.5 &  33.5 $\pm$ 2.0 &  40.2 $\pm$ 1.2 &  45.2 $\pm$ 0.5 &   47.7 $\pm$ 0.0 &              &              \\
        SpineWeb &  28.1 $\pm$ 2.1 &  39.3 $\pm$ 6.1 &  52.1 $\pm$ 7.0 &  63.1 $\pm$ 3.3 &   64.7 $\pm$ 0.0 &              &              \\
      WBC\textsubscript{0} &  49.4 $\pm$ 4.5 &  65.0 $\pm$ 4.3 &  74.8 $\pm$ 3.0 &  81.0 $\pm$ 1.9 &   85.0 $\pm$ 1.2 &   87.5 $\pm$ 0.8 &   88.8 $\pm$ 0.0 \\
      WBC\textsubscript{1} &  57.4 $\pm$ 4.8 &  75.2 $\pm$ 3.5 &  83.0 $\pm$ 2.1 &  87.4 $\pm$ 1.0 &   89.9 $\pm$ 0.4 &   91.3 $\pm$ 0.3 &   91.9 $\pm$ 0.2 \\
\bottomrule
\end{tabular}

}
\label{tab:inference-size-per-task}
\end{table}

\begin{table}
\centering
\rowcolors{2}{gray!15}{white}
\caption{\textbf{Ensembling predictions at different inference support sizes.}
For each inference support size $N$, we report the results (in average held-out Dice Score) of taking 100 predictions ($K=1$) and ensembling by averaging in groups of size $K$, performing 100 repetitions for each $K$.
We report the mean and standard deviation across the 100 values for each setting and find that increasing either $K$ or $N$ leads to improved model performance, with $N$ having a significantly larger effect than $K$.
}
\begin{tabular}{rrrrrrrr}
 \textbf{N} & \textbf{K = 1} & \textbf{K = 2} & \textbf{K = 4} & \textbf{K = 8} & \textbf{K = 16} & \textbf{K = 32} & \textbf{K = 64} \\
\toprule
       1 &  36.9 $\pm$ 2.0 &  39.4 $\pm$ 2.3 &  40.7 $\pm$ 2.0 &  40.9 $\pm$ 1.6 &   40.3 $\pm$ 1.1 &   39.4 $\pm$ 0.7 &   38.3 $\pm$ 0.4 \\
       2 &  51.0 $\pm$ 3.2 &  56.3 $\pm$ 2.2 &  59.5 $\pm$ 1.6 &  61.0 $\pm$ 1.2 &   61.9 $\pm$ 0.9 &   62.3 $\pm$ 0.6 &   62.4 $\pm$ 0.3 \\
       4 &  59.4 $\pm$ 2.3 &  63.7 $\pm$ 1.5 &  66.2 $\pm$ 1.0 &  67.5 $\pm$ 0.7 &   68.2 $\pm$ 0.4 &   68.6 $\pm$ 0.3 &   68.8 $\pm$ 0.2 \\
       8 &  64.8 $\pm$ 1.9 &  68.0 $\pm$ 1.1 &  69.6 $\pm$ 0.6 &  70.5 $\pm$ 0.4 &   71.1 $\pm$ 0.2 &   71.3 $\pm$ 0.2 &   71.4 $\pm$ 0.1 \\
      16 &  68.4 $\pm$ 1.1 &  70.1 $\pm$ 0.5 &  71.0 $\pm$ 0.4 &  71.5 $\pm$ 0.3 &   71.8 $\pm$ 0.2 &   71.9 $\pm$ 0.1 &   72.0 $\pm$ 0.1 \\
      32 &  70.1 $\pm$ 0.6 &  71.0 $\pm$ 0.3 &  71.5 $\pm$ 0.2 &  71.7 $\pm$ 0.1 &   71.8 $\pm$ 0.1 &   71.9 $\pm$ 0.1 &   71.9 $\pm$ 0.0 \\
      64 &  71.0 $\pm$ 0.3 &  71.4 $\pm$ 0.2 &  71.6 $\pm$ 0.2 &  71.7 $\pm$ 0.1 &   71.8 $\pm$ 0.1 &   71.8 $\pm$ 0.1 &   71.8 $\pm$ 0.0 \\
\bottomrule
\end{tabular}

\label{tab:ensemble-results}
\end{table}

\begin{table}
\centering
\caption{
\textbf{Number of Training Datasets and Tasks}.
Test Dice score results for models trained with N\textsubscript{D} datasets comprising  N\textsubscript{T} tasks.
The subsets of the training datasets are chosen randomly so we report three realizations for each  N\textsubscript{D}, except for the case where all datasets are included.
Each row corresponds to a separate \method{} model.
}
{\renewcommand{\arraystretch}{1.2}
\begin{tabular}{rrrrrrrrr}
\textbf{N\textsubscript{D}} & \textbf{N\textsubscript{T}} &  \textbf{ACDC} & \textbf{PanDental} &   \textbf{SCD} & \textbf{STARE} & \textbf{SpineWeb} &   \textbf{WBC} & \textbf{All (avg)} \\
\toprule
\multirow{3}{*}{1} & 25   &  24.7 $\pm$ 2.8 &      82.2 $\pm$ 0.8 &  43.7 $\pm$ 3.3 &   7.2 $\pm$ 2.5 &      0.2 $\pm$ 0.2 &  61.1 $\pm$ 1.3 &       36.5 $\pm$ 0.8 \\
   & 29   &   3.3 $\pm$ 2.4 &      18.4 $\pm$ 0.7 &   0.2 $\pm$ 0.1 &   0.0 $\pm$ 0.0 &      0.0 $\pm$ 0.0 &   0.0 $\pm$ 0.0 &        3.7 $\pm$ 0.4 \\
   & 156  &  63.1 $\pm$ 2.7 &      80.0 $\pm$ 1.8 &  53.5 $\pm$ 5.3 &  17.5 $\pm$ 2.3 &     46.2 $\pm$ 1.6 &  86.2 $\pm$ 1.3 &       57.8 $\pm$ 1.1 \\
\cline{1-9}
\multirow{3}{*}{2} & 33   &  41.7 $\pm$ 3.7 &      83.8 $\pm$ 1.0 &  43.9 $\pm$ 2.3 &  16.9 $\pm$ 1.4 &     22.2 $\pm$ 3.3 &  80.0 $\pm$ 1.5 &       48.1 $\pm$ 1.1 \\
   & 85   &  49.5 $\pm$ 3.5 &      83.1 $\pm$ 1.1 &  59.9 $\pm$ 2.5 &  22.9 $\pm$ 2.1 &     52.9 $\pm$ 2.1 &  85.5 $\pm$ 1.4 &       59.0 $\pm$ 1.0 \\
   & 131  &  31.9 $\pm$ 2.5 &      82.0 $\pm$ 0.8 &  42.3 $\pm$ 3.4 &   0.0 $\pm$ 0.0 &     10.2 $\pm$ 3.2 &  69.5 $\pm$ 1.3 &       39.3 $\pm$ 0.6 \\
\cline{1-9}
\multirow{3}{*}{5} & 237  &  43.6 $\pm$ 5.5 &      75.3 $\pm$ 1.3 &  52.1 $\pm$ 2.8 &  26.8 $\pm$ 3.4 &    28.3 $\pm$ 10.1 &  86.3 $\pm$ 1.2 &       52.1 $\pm$ 2.2 \\
   & 710  &  63.9 $\pm$ 3.6 &      86.6 $\pm$ 1.0 &  63.5 $\pm$ 2.2 &  27.2 $\pm$ 3.3 &     61.1 $\pm$ 5.0 &  87.4 $\pm$ 1.6 &       64.9 $\pm$ 1.3 \\
   & 1117 &  67.4 $\pm$ 3.1 &      86.5 $\pm$ 1.0 &  62.9 $\pm$ 4.5 &  51.6 $\pm$ 2.4 &     58.4 $\pm$ 8.2 &  89.9 $\pm$ 1.0 &       69.4 $\pm$ 2.0 \\
\cline{1-9}
\multirow{3}{*}{11} & 174  &  51.2 $\pm$ 2.8 &      84.0 $\pm$ 0.9 &  66.5 $\pm$ 2.7 &  22.8 $\pm$ 1.3 &     52.2 $\pm$ 0.4 &  84.5 $\pm$ 1.1 &       60.2 $\pm$ 0.8 \\
   & 1223 &  60.6 $\pm$ 3.6 &      86.8 $\pm$ 1.0 &  59.0 $\pm$ 4.9 &  29.7 $\pm$ 2.4 &     58.4 $\pm$ 5.6 &  85.5 $\pm$ 1.3 &       63.3 $\pm$ 1.7 \\
   & 1457 &  69.7 $\pm$ 2.8 &      87.5 $\pm$ 0.9 &  65.4 $\pm$ 3.3 &  40.5 $\pm$ 3.0 &     59.6 $\pm$ 7.0 &  88.6 $\pm$ 1.1 &       68.6 $\pm$ 1.7 \\
\cline{1-9}
\multirow{3}{*}{23} & 1320 &  66.9 $\pm$ 3.5 &      85.3 $\pm$ 0.9 &  68.1 $\pm$ 2.4 &  31.2 $\pm$ 0.7 &     57.2 $\pm$ 5.9 &  88.4 $\pm$ 1.3 &       66.2 $\pm$ 1.1 \\
   & 2157 &  66.5 $\pm$ 3.0 &      86.1 $\pm$ 1.0 &  64.0 $\pm$ 2.3 &  36.9 $\pm$ 1.1 &     64.9 $\pm$ 5.4 &  89.6 $\pm$ 1.3 &       68.0 $\pm$ 1.1 \\
   & 2276 &  66.5 $\pm$ 3.8 &      86.7 $\pm$ 0.9 &  65.7 $\pm$ 2.8 &  28.1 $\pm$ 2.4 &     52.4 $\pm$ 6.6 &  89.1 $\pm$ 1.2 &       64.8 $\pm$ 1.4 \\
\cline{1-9}
\multirow{3}{*}{34} & 3008 &  69.8 $\pm$ 3.0 &      88.8 $\pm$ 0.9 &  68.3 $\pm$ 2.6 &  42.5 $\pm$ 3.2 &     62.4 $\pm$ 6.1 &  90.0 $\pm$ 1.2 &       70.3 $\pm$ 1.4 \\
   & 3483 &  70.7 $\pm$ 2.9 &      87.1 $\pm$ 1.0 &  67.2 $\pm$ 3.5 &  46.7 $\pm$ 3.6 &     63.0 $\pm$ 5.0 &  90.7 $\pm$ 1.4 &       70.9 $\pm$ 1.0 \\
   & 3854 &  65.3 $\pm$ 3.6 &      88.2 $\pm$ 1.0 &  65.1 $\pm$ 1.6 &  43.1 $\pm$ 3.1 &     62.2 $\pm$ 5.5 &  89.0 $\pm$ 1.1 &       68.8 $\pm$ 1.0 \\
\cline{1-9}
46 & 4432 &  71.3 $\pm$ 2.6 &      87.9 $\pm$ 0.9 &  67.9 $\pm$ 2.5 &  44.9 $\pm$ 2.9 &     65.5 $\pm$ 4.7 &  91.0 $\pm$ 1.1 &       71.4 $\pm$ 1.1 \\
\bottomrule
\end{tabular}

}
\label{tab:numtasks}
\end{table}

\clearpage
\subsection{Additional Results}

\subpara{Few-shot Baseline Model Variants}.
The FS baselines (ALPNet, PANet, and SENet) were introduced in a few-shot setting where the underlying assumption is that any new task can only have very few examples, rather than our setting where we avoid re-training due to the limitations of the clinical settings. These baselines were therefore presented with a support size of 1 example. They also involved no data or task augmentation.
Our ablations show that \method{} models performed best with large support set sizes and increased data and task diversity from augmenting examples.
Consequently, we test whether incorporating these changes to the baseline methods leads to improved performance in the held-out datasets in our setting, where more data *might* be available for some datasets.
Similarly, we also test whether ensembling predictions from several support sets lead to better predictions, as we do for \method{}.

Table~\ref{tab:fewshot-hyperparam} and Table~\ref{tab:fewshot-hyperparam-per-dataset} report results of the hyperparameter grid search for all the few-shot baseline models and \method{}.
Table~\ref{tab:fewshot-hyperparam} shows that ensembling ($K=10$) and an increased support size ($N=64$) leads to held-out improvements for all methods.
In contrast, augmentation strategies do not benefit all methods. While \method{} and SENet improve when using augmentation strategies, PANet and ALPNet experience a decrease in performance.
Table~\ref{tab:fewshot-hyperparam-per-dataset} shows that the best hyperparameter setting is not consistent across held-out datasets for the baseline methods.

\begin{table}[!h]
\centering
\caption{
\textbf{FS baseline hyperparameter search}.
For each method, we report results for models trained with a support size $N$, ensemble size $K$, and with and without data and task augmentation.
Dice scores are averaged across all datasets and the standard deviation is computed via subject-level bootstrapping.
}
{
\setlength{\cmidrulewidth}{0.9pt}
\begin{tabular}{lrcccc}
                 && \multicolumn{2}{c}{\textbf{No Aug}} & \multicolumn{2}{c}{\textbf{Aug}} \\
\cmidrule(lr){3-6}
                \textbf{Model} & \textbf{N} &               K=1 &              K=10 &               K=1 &              K=10 \\
Model & N &                   &                   &                   &                   \\
\toprule
\multirow{2}{*}{ALPNet} & 1  &        40.2 $\pm$ 0.9 &        42.3 $\pm$ 1.3 &        35.4 $\pm$ 0.6 &        37.0 $\pm$ 0.8 \\
                 & 64 &        46.3 $\pm$ 1.3 &  \textbf{47.8 $\pm$ 1.1} &        42.3 $\pm$ 1.0 &        45.2 $\pm$ 1.2 \\
\cline{1-6}
\multirow{2}{*}{PANet} & 1  &        37.4 $\pm$ 0.7 &        39.3 $\pm$ 0.8 &        33.2 $\pm$ 1.3 &        34.3 $\pm$ 1.4 \\
                 & 64 &  \textbf{41.6 $\pm$ 1.3} &  \textbf{41.8 $\pm$ 1.3} &        38.7 $\pm$ 0.9 &        40.8 $\pm$ 0.8 \\
\cline{1-6}
\multirow{2}{*}{SENet} & 1  &        40.0 $\pm$ 0.9 &        41.2 $\pm$ 0.9 &        40.1 $\pm$ 1.2 &        41.1 $\pm$ 1.4 \\
                 & 64 &        42.1 $\pm$ 0.7 &        42.4 $\pm$ 0.8 &  \textbf{50.2 $\pm$ 1.1} &  \textbf{50.1 $\pm$ 1.3} \\
\cline{1-6}
\multirow{2}{*}{UniverSeg (ours)} & 1  &        49.7 $\pm$ 0.9 &        53.4 $\pm$ 1.1 &        51.9 $\pm$ 0.8 &        54.0 $\pm$ 1.0 \\
                 & 64 &        64.0 $\pm$ 1.1 &        64.5 $\pm$ 1.0 &        71.0 $\pm$ 1.0 &  \textbf{71.8 $\pm$ 0.9} \\
\bottomrule
\end{tabular}

}
\label{tab:fewshot-hyperparam}
\end{table}

\begin{table}[!h]

\centering
\caption{
\textbf{Few-shot baseline hyperparameter search per dataset}.
For each method, we report results for models trained with a support size $N$, ensemble size $K=10$, and with and without data and task augmentation.
Dice score values are averaged across all datasets and the standard deviation is computed via subject-level bootstrapping.
For each dataset and model, we highlight the setting with the best performance
}
{\renewcommand{\arraystretch}{1.2}
\begin{tabular}{lclcccccc}
\textbf{Model} & \textbf{N} & \textbf{Aug} &        \textbf{ACDC} &   \textbf{PanDental} &         \textbf{SCD} &       \textbf{STARE} &    \textbf{SpineWeb} &         \textbf{WBC} \\
\toprule
\multirow{4}{*}{ALPNet} & \multirow{2}{*}{1} & No &        22.1 $\pm$ 3.2 &        66.8 $\pm$ 1.0 &        49.1 $\pm$ 3.8 &  \textbf{22.7 $\pm$ 2.0} &        29.7 $\pm$ 3.7 &        63.2 $\pm$ 0.9 \\
          &    & Yes &        26.7 $\pm$ 2.9 &        51.8 $\pm$ 1.5 &        41.5 $\pm$ 1.9 &        11.0 $\pm$ 2.8 &        19.7 $\pm$ 4.9 &        71.0 $\pm$ 1.6 \\
\cline{2-9}
          & \multirow{2}{*}{64} & No &        34.6 $\pm$ 2.4 &  \textbf{72.9 $\pm$ 0.8} &        53.4 $\pm$ 3.0 &        17.8 $\pm$ 1.9 &  \textbf{31.6 $\pm$ 4.6} &  \textbf{76.2 $\pm$ 1.1} \\
          &    & Yes &  \textbf{38.3 $\pm$ 2.5} &        71.1 $\pm$ 1.0 &  \textbf{56.1 $\pm$ 1.6} &         6.3 $\pm$ 2.2 &        25.5 $\pm$ 6.4 &        73.9 $\pm$ 1.2 \\
\midrule 
 \midrule 
\multirow{4}{*}{PANet} & \multirow{2}{*}{1} & No &  \textbf{33.4 $\pm$ 2.5} &  \textbf{69.8 $\pm$ 1.3} &        48.7 $\pm$ 3.5 &        17.4 $\pm$ 4.3 &        25.4 $\pm$ 3.9 &        40.9 $\pm$ 1.8 \\
          &    & Yes &        30.3 $\pm$ 3.0 &        63.2 $\pm$ 1.3 &        48.4 $\pm$ 3.3 &         4.6 $\pm$ 2.9 &  \textbf{28.6 $\pm$ 5.6} &        31.0 $\pm$ 2.1 \\
\cline{2-9}
          & \multirow{2}{*}{64} & No &        27.8 $\pm$ 4.3 &        67.7 $\pm$ 0.8 &  \textbf{58.9 $\pm$ 3.4} &  \textbf{20.1 $\pm$ 3.2} &        21.8 $\pm$ 0.4 &        54.7 $\pm$ 1.6 \\
          &    & Yes &        29.6 $\pm$ 2.3 &        66.4 $\pm$ 1.4 &        46.8 $\pm$ 2.3 &        15.1 $\pm$ 2.1 &        27.9 $\pm$ 5.8 &  \textbf{58.8 $\pm$ 1.5} \\
\midrule 
 \midrule 
\multirow{4}{*}{SENet} & \multirow{2}{*}{1} & No &        17.0 $\pm$ 2.9 &        61.7 $\pm$ 1.1 &        47.5 $\pm$ 2.3 &  \textbf{41.3 $\pm$ 2.7} &  \textbf{21.7 $\pm$ 3.7} &        58.1 $\pm$ 0.9 \\
          &    & Yes &        32.2 $\pm$ 2.8 &        62.4 $\pm$ 1.3 &        48.2 $\pm$ 2.9 &        31.4 $\pm$ 2.1 &        16.8 $\pm$ 7.8 &        55.5 $\pm$ 1.3 \\
\cline{2-9}
          & \multirow{2}{*}{64} & No &        32.0 $\pm$ 2.4 &        79.1 $\pm$ 0.8 &        43.8 $\pm$ 2.9 &        37.5 $\pm$ 2.9 &         3.2 $\pm$ 2.4 &        58.7 $\pm$ 1.1 \\
          &    & Yes &  \textbf{40.1 $\pm$ 2.0} &  \textbf{81.1 $\pm$ 0.9} &  \textbf{55.4 $\pm$ 3.3} &        35.2 $\pm$ 2.2 &        18.3 $\pm$ 4.0 &  \textbf{70.8 $\pm$ 1.3} \\
\midrule 
 \midrule 
\multirow{4}{*}{UniverSeg} & \multirow{2}{*}{1} & No &        29.1 $\pm$ 2.0 &        76.1 $\pm$ 0.9 &        58.0 $\pm$ 2.1 &        54.5 $\pm$ 2.7 &        31.6 $\pm$ 6.4 &        70.9 $\pm$ 1.7 \\
          &    & Yes &        37.5 $\pm$ 2.0 &        76.8 $\pm$ 1.1 &        62.9 $\pm$ 2.6 &        33.9 $\pm$ 3.7 &        36.0 $\pm$ 5.0 &        76.8 $\pm$ 1.6 \\
\cline{2-9}
          & \multirow{2}{*}{64} & No &        50.6 $\pm$ 2.9 &        85.7 $\pm$ 0.9 &        59.0 $\pm$ 1.9 &  \textbf{61.9 $\pm$ 1.6} &        45.6 $\pm$ 4.8 &        84.2 $\pm$ 1.4 \\
          &    & Yes &  \textbf{70.9 $\pm$ 2.9} &  \textbf{87.5 $\pm$ 0.9} &  \textbf{69.0 $\pm$ 2.9} &        48.1 $\pm$ 2.0 &  \textbf{64.6 $\pm$ 5.4} &  \textbf{90.6 $\pm$ 1.1} \\
\bottomrule
\end{tabular}

}
\label{tab:fewshot-hyperparam-per-dataset}
\end{table}

\clearpage

\begin{figure*}[!h]
\centerline{\includegraphics[width=0.74\textwidth]{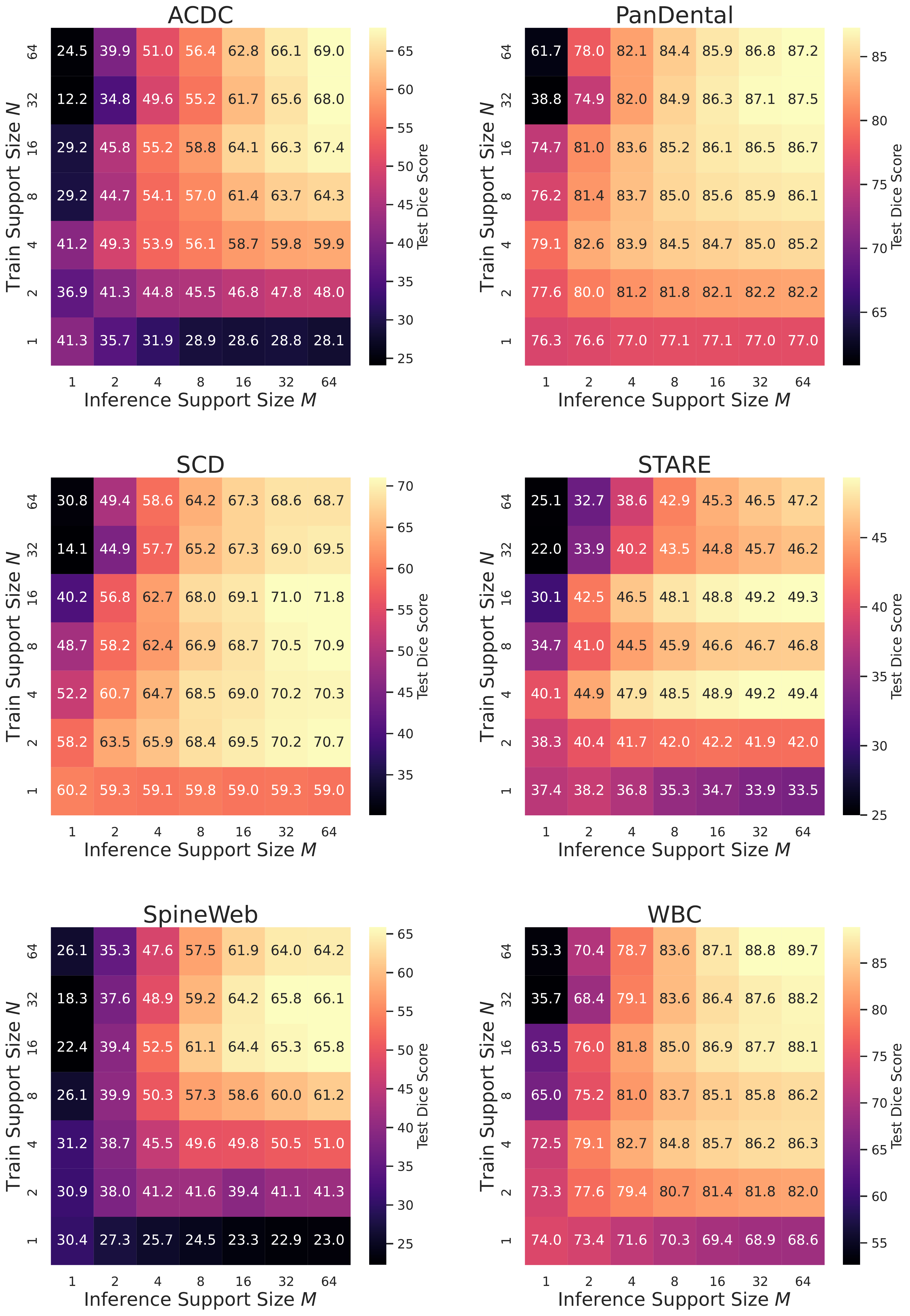}}
\caption{%
\textbf{Cartesian Product of Training and Inference Support Sizes.}
Test results for using a \method{} trained with a support size of $N$ examples and performing inference with a support size of $M$ examples.
No ensembling is performed ($K=1$), but we perform 10 repetitions with varying support sets and report the average.
}
\label{fig:SupportSizeHeatmap}
\end{figure*}

\subpara{Using different training and inference support sizes}.
In Figure~\ref{fig:SupportSizeHeatmap}, we report dataset-level results of performing inference with a support size of $M$ using a \method{} model trained with a support size of $N$ examples.
We find that using support sets larger than those seen in training ($M>N$, lower quadrant of heat-maps) leads to improvements for $N \geq 2$, which demonstrates the model is learning to interact the elements of the support set and benefits from larger amounts of examples.

\clearpage

\section{Additional Visualizations}
\label{sec: supp-ext-results}

\begin{figure*}[!h]
\centerline{\includegraphics[width=\textwidth]{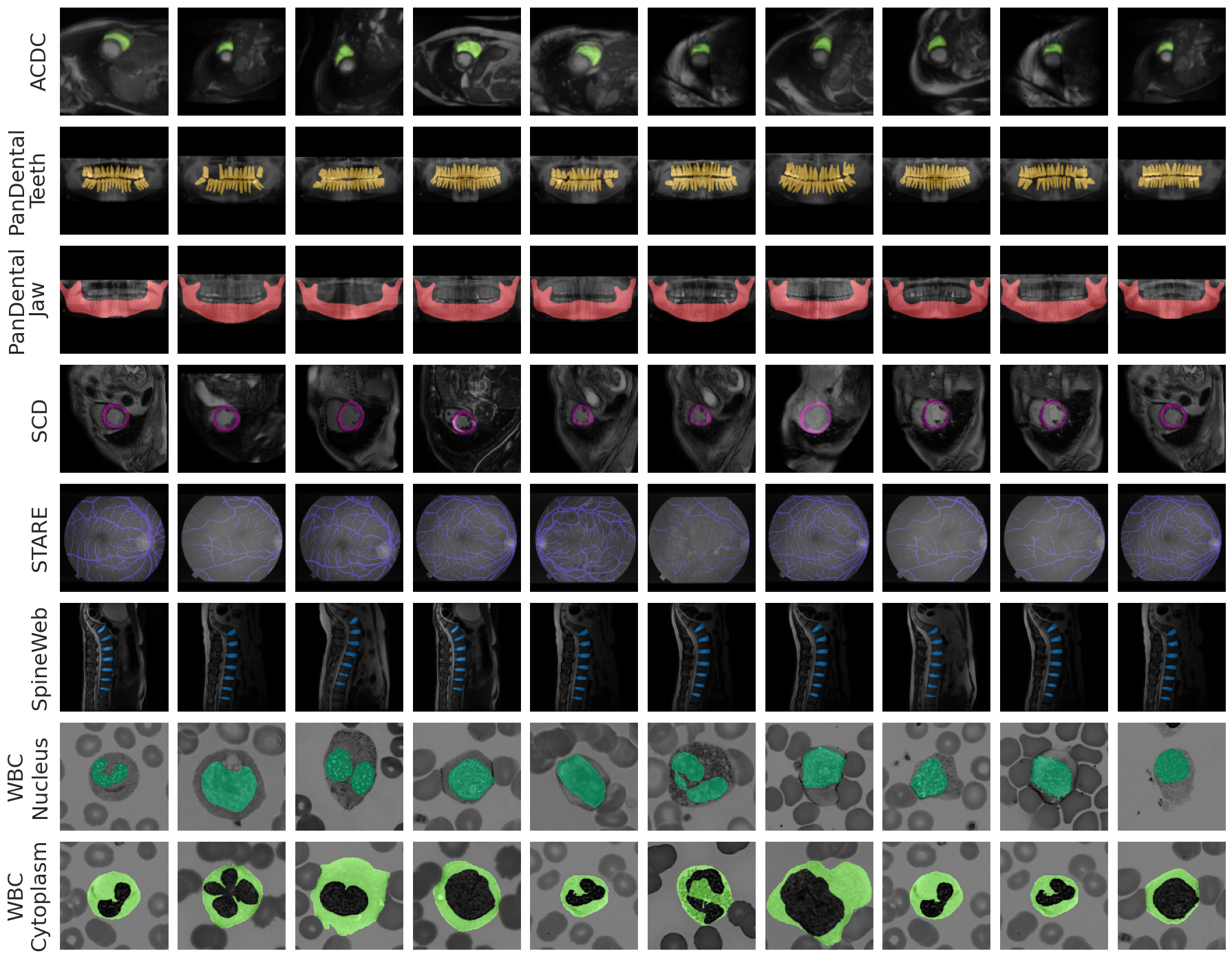}}
\caption{\textbf{Example Support Sets for Held-Out Datasets.}}
\label{fig:SupportExamples}
\end{figure*}

\subpara{Visualization of Held-Out Support Sets.} \method{} networks take advantage of large support sets of (image, label-map) pairs, which can be very diverse. In Figure~\ref{fig:SupportExamples}, we visualize a random subset of 10 pairs for each held-out dataset. The diversity of subjects amongst support sets differs between tasks, which likely plays a role in the number of examples required to perform well.

\begin{figure*}[!h]
\centerline{\includegraphics[width=\textwidth]{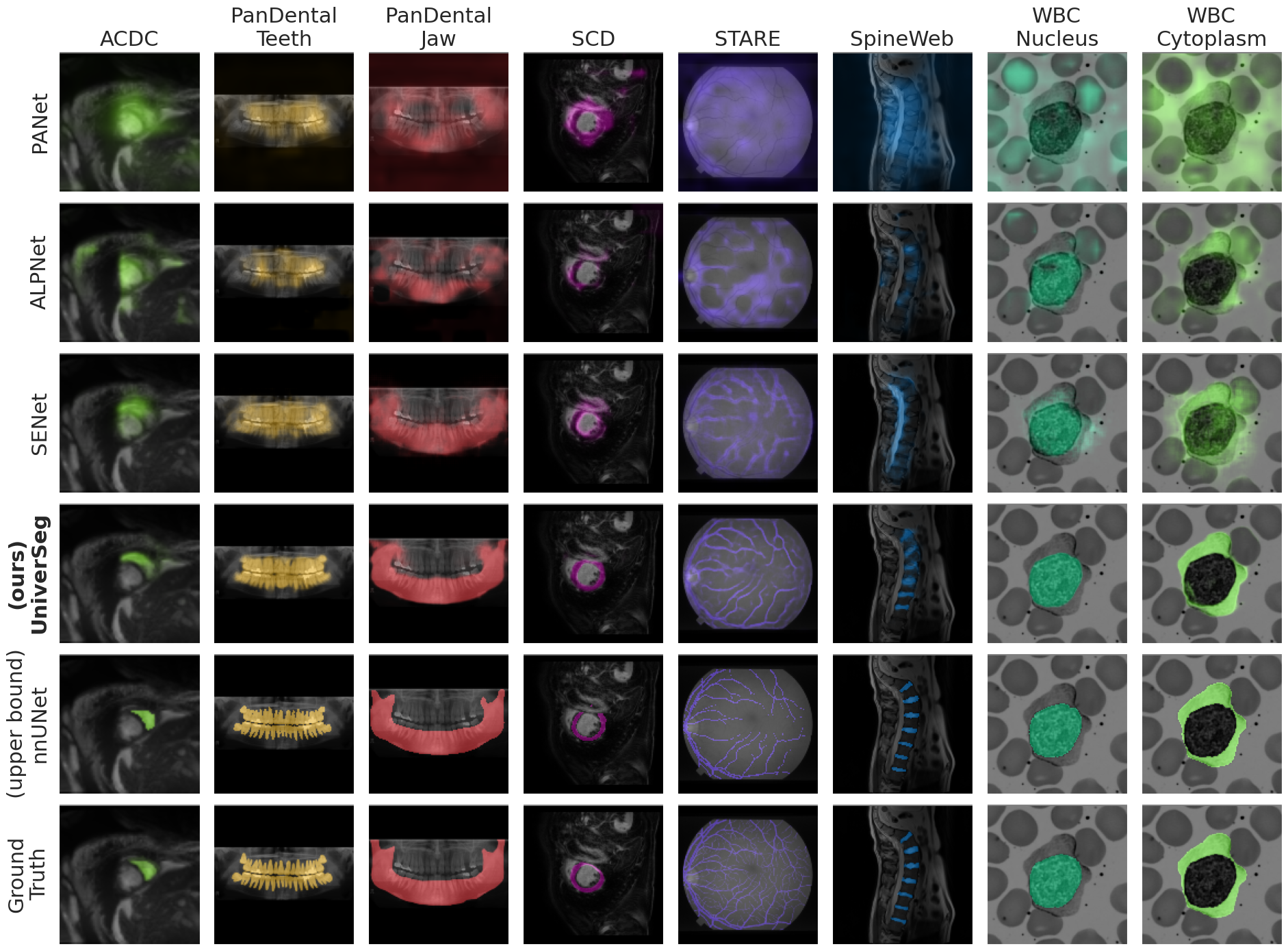}}
\caption{\textbf{Visualization of Soft (Non-Thresholded) Predictions for All Methods.}}
\label{fig:SoftPredictions}
\end{figure*}

\clearpage
\subpara{Visualization of Soft Predictions.} Thresholding segmentation predictions (at 0.5) provides a binary segmentation and enables computation of well-known metrics such as the (hard) Dice score. However, for certain regions of interest, like thin structures, thresholding can hide network performance. In Figure~\ref{fig:SoftPredictions}, we show this effect visually. For example, focusing on STARE, we see that \method{} networks can capture the thin structures very well, which is lost when thresholding the predictions to create a binary segmentation.

\clearpage
\subpara{WBC task visualizations}.
We include some visualizations of \method{}'s capability to adapt based on the support set specification. We use the WBC dataset, which presents substantial variability between support set examples.

\begin{figure*}[!h]
\begin{subfigure}{\textwidth}
\centering
\caption{\textbf{Support Set Examples - Cytoplasm Label}}
\includegraphics[width=0.8\textwidth]{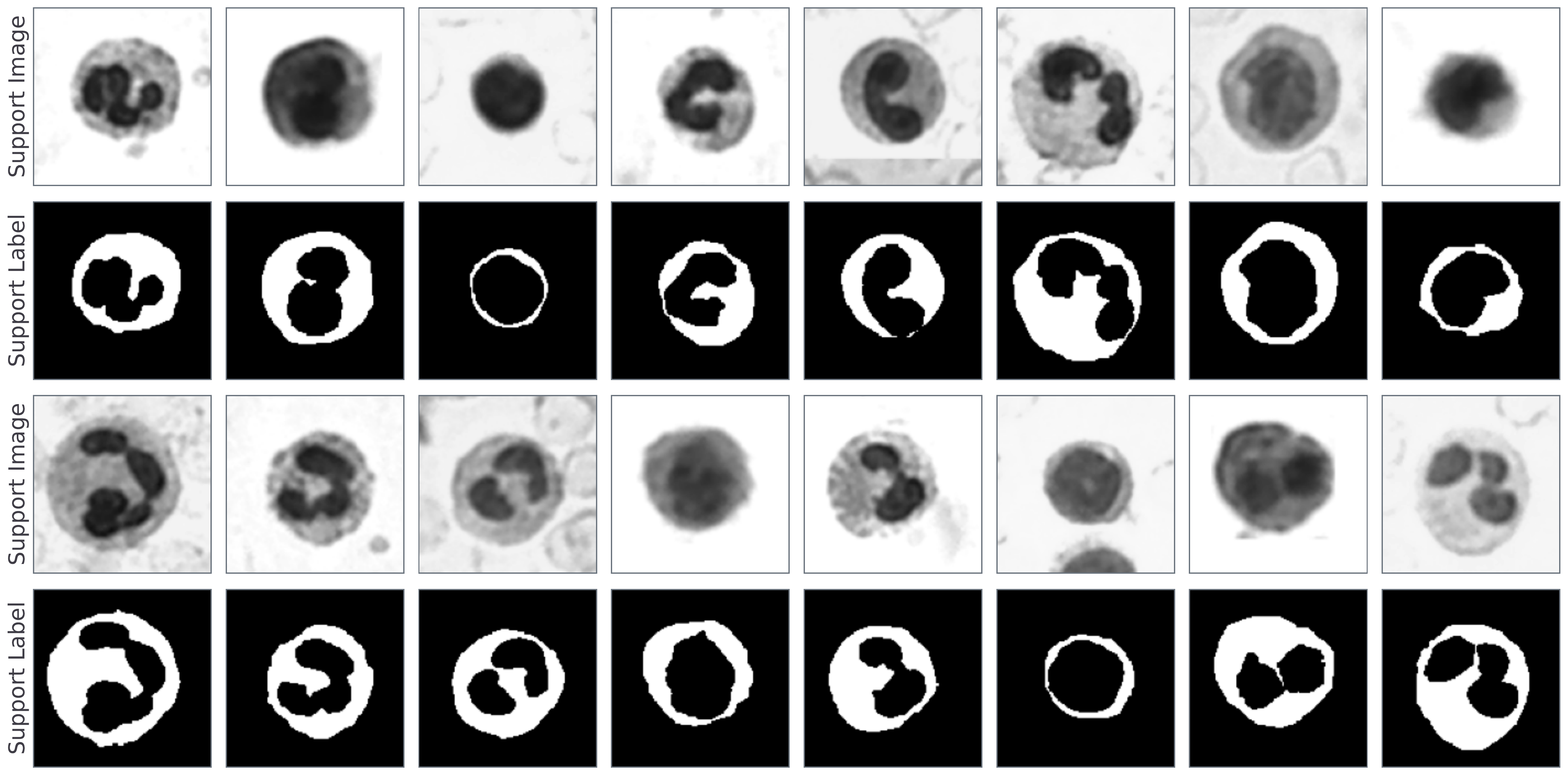}
\end{subfigure}
\begin{subfigure}{\textwidth}
\caption{\textbf{Predictions - Cytoplasm Label}}
\includegraphics[width=\textwidth]{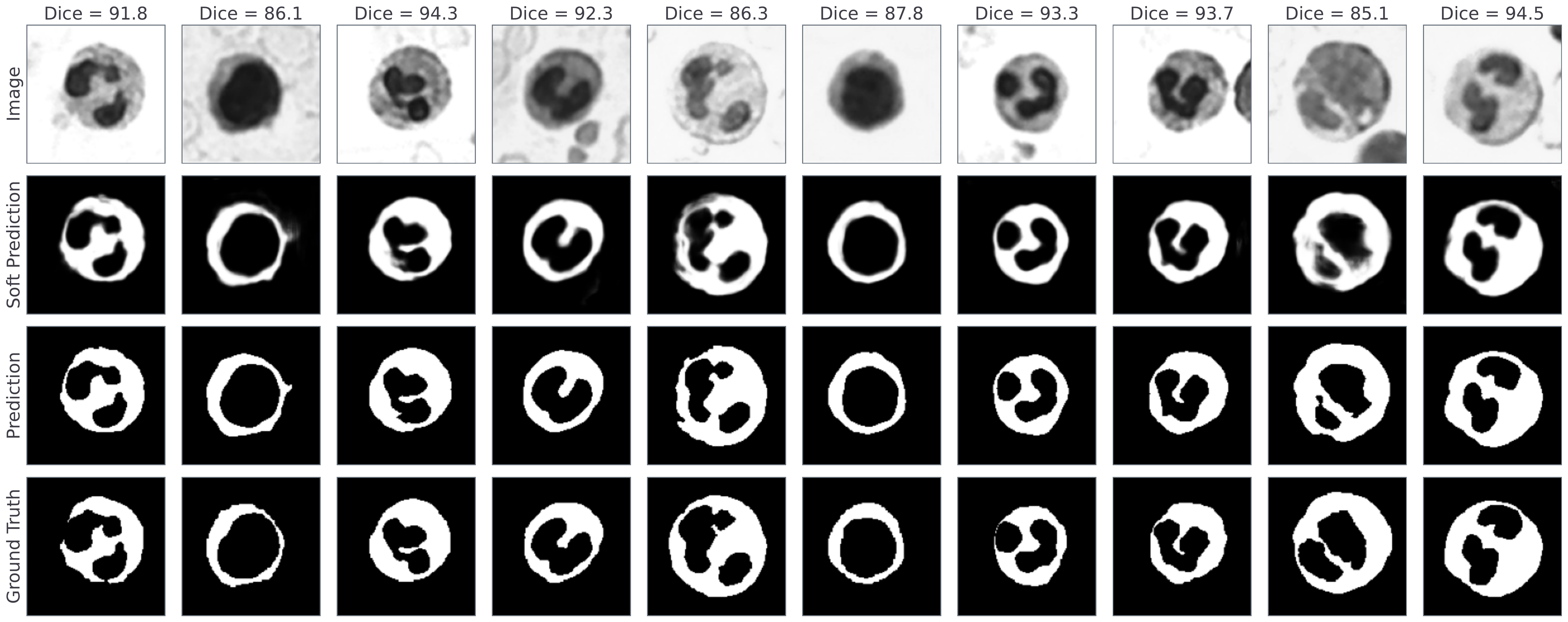}
\end{subfigure}
\caption{Visualization of support set examples (a) and predictions (b) for the WBC Cytoplasm label}
\label{fig:wbc_viz_01}
\end{figure*}

\begin{figure*}[!h]
\begin{subfigure}{\textwidth}
\centering
\caption{\textbf{Support Set Examples - Nucleus Label}}
\includegraphics[width=0.8\textwidth]{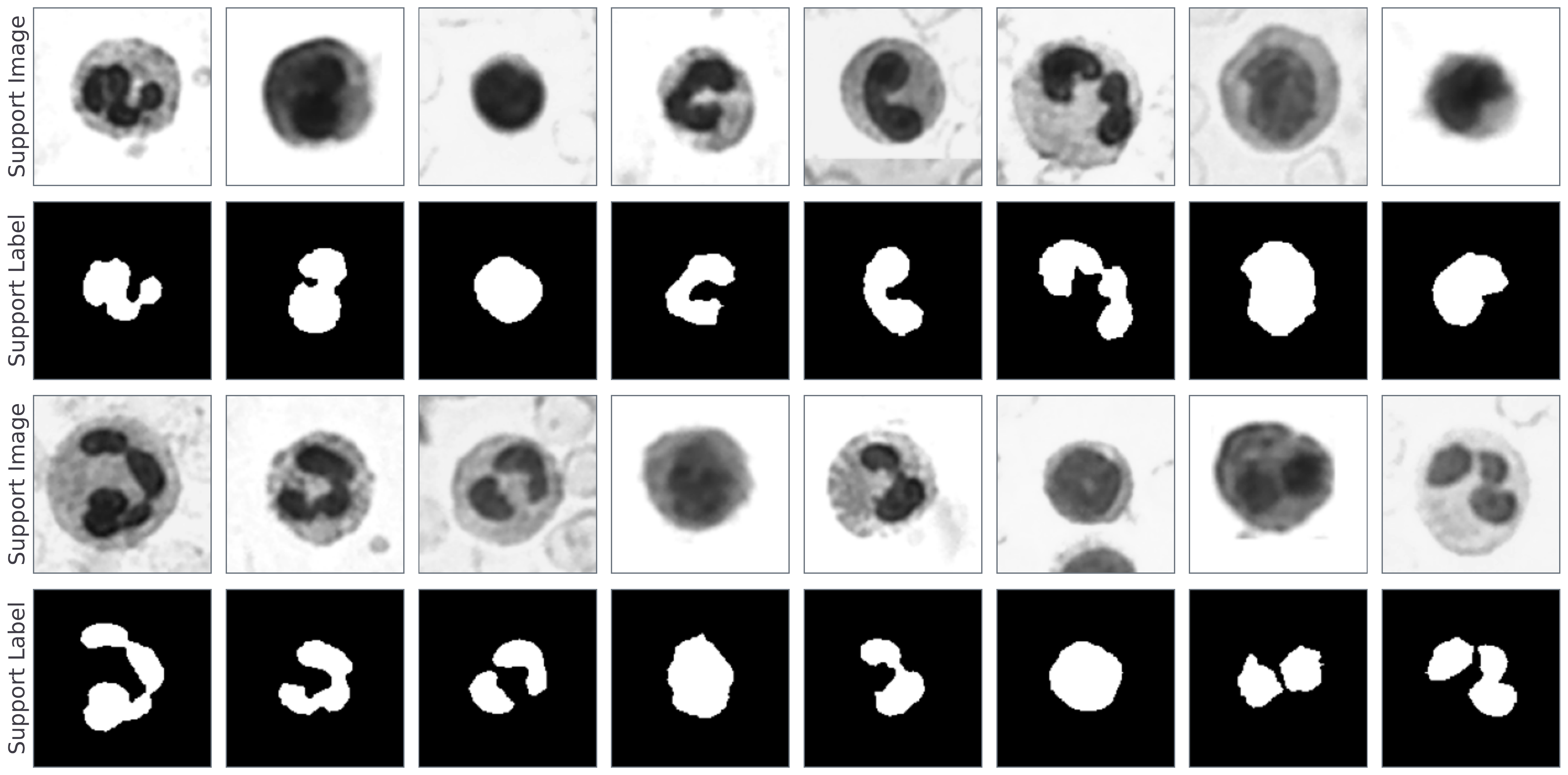}
\end{subfigure}
\begin{subfigure}{\textwidth}
\caption{\textbf{Predictions - Nucleus Label}}
\includegraphics[width=\textwidth]{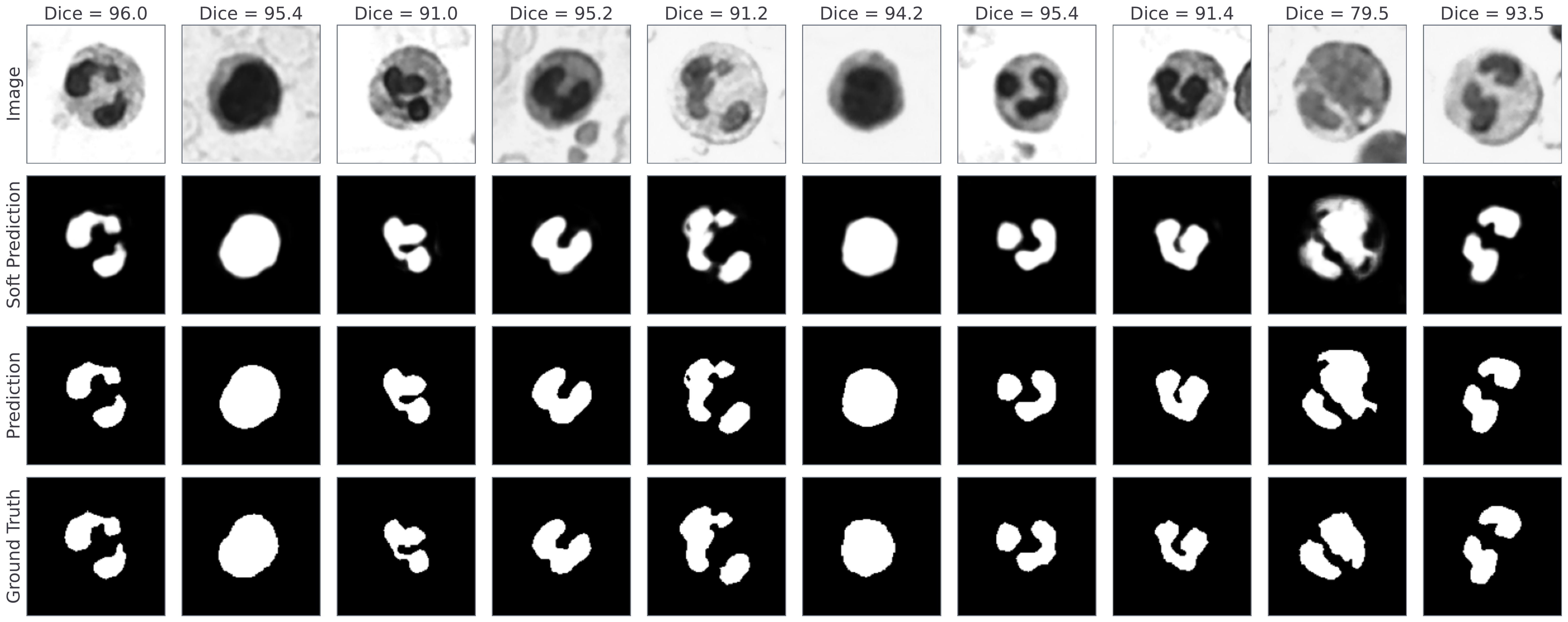}
\end{subfigure}
\caption{Visualization of support set examples (a) and predictions (b) for the WBC Nucleus label}
\label{fig:wbc_viz_02}
\end{figure*}

\begin{figure*}[!h]
\begin{subfigure}{\textwidth}
\centering
\caption{\textbf{Support Set Examples - Inverted Images}}
\includegraphics[width=0.8\textwidth]{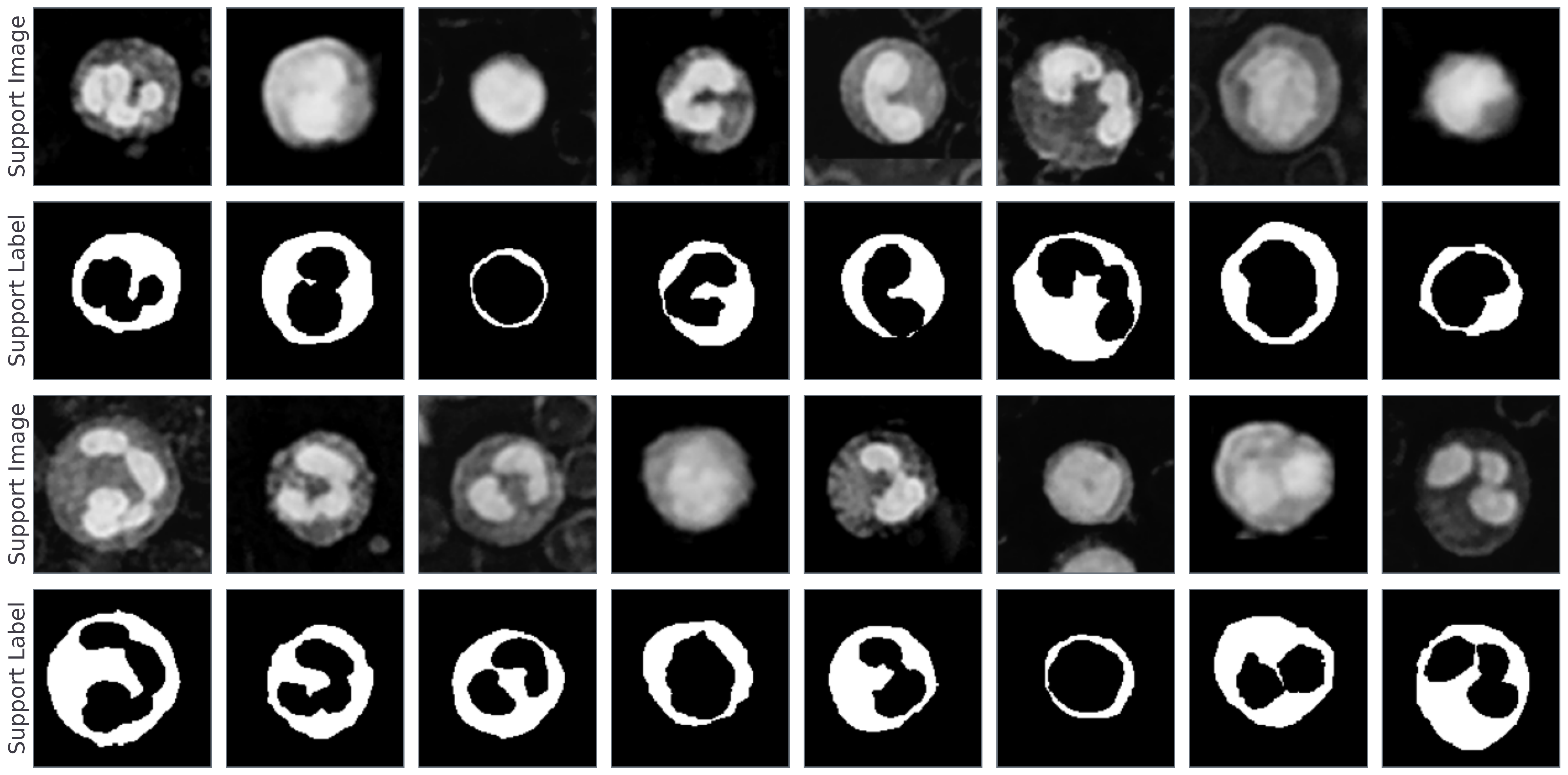}
\end{subfigure}
\begin{subfigure}{\textwidth}
\caption{\textbf{Predictions - Inverted Images}}
\includegraphics[width=\textwidth]{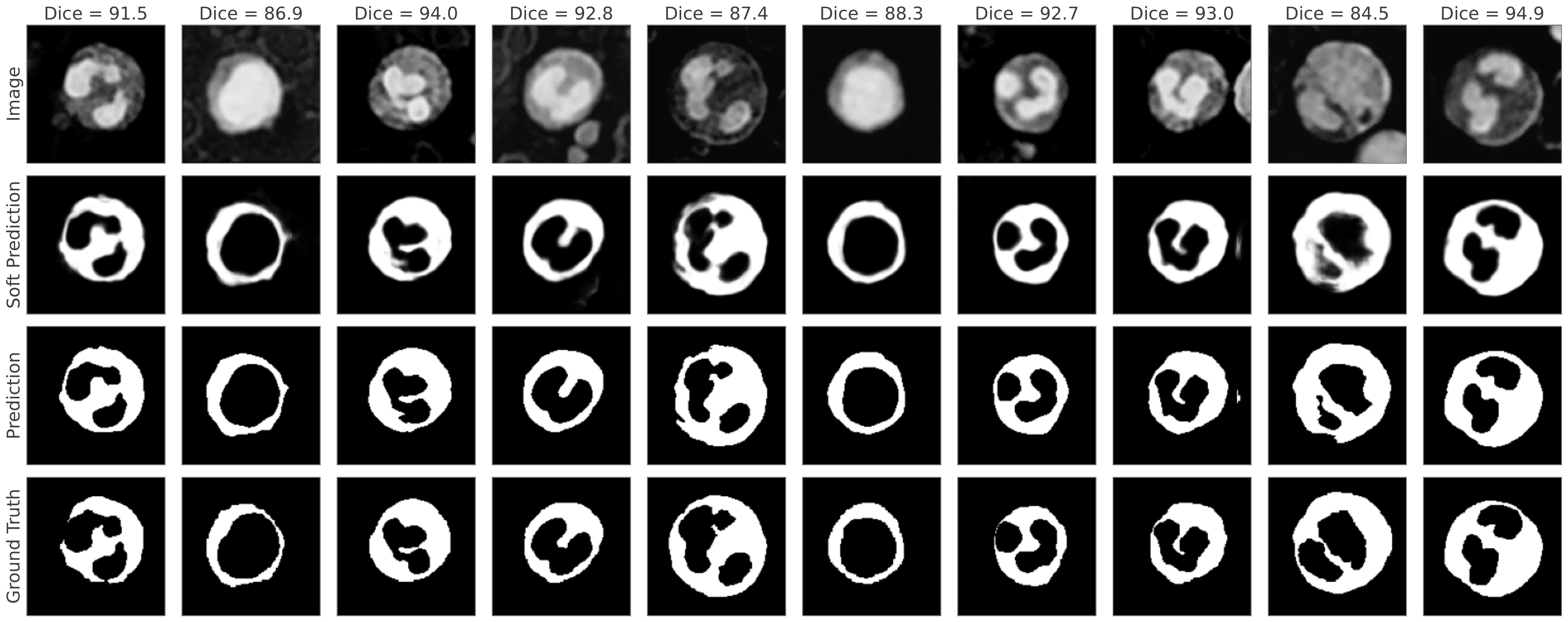}
\end{subfigure}
\caption{Visualization of support set examples (a) and predictions (b) for the WBC Cytoplasm label with inverted images}
\label{fig:wbc_viz_03}
\end{figure*}

\begin{figure*}[!h]
\begin{subfigure}{\textwidth}
\centering
\caption{\textbf{Support Set Examples - Multi Label}}
\includegraphics[width=0.8\textwidth]{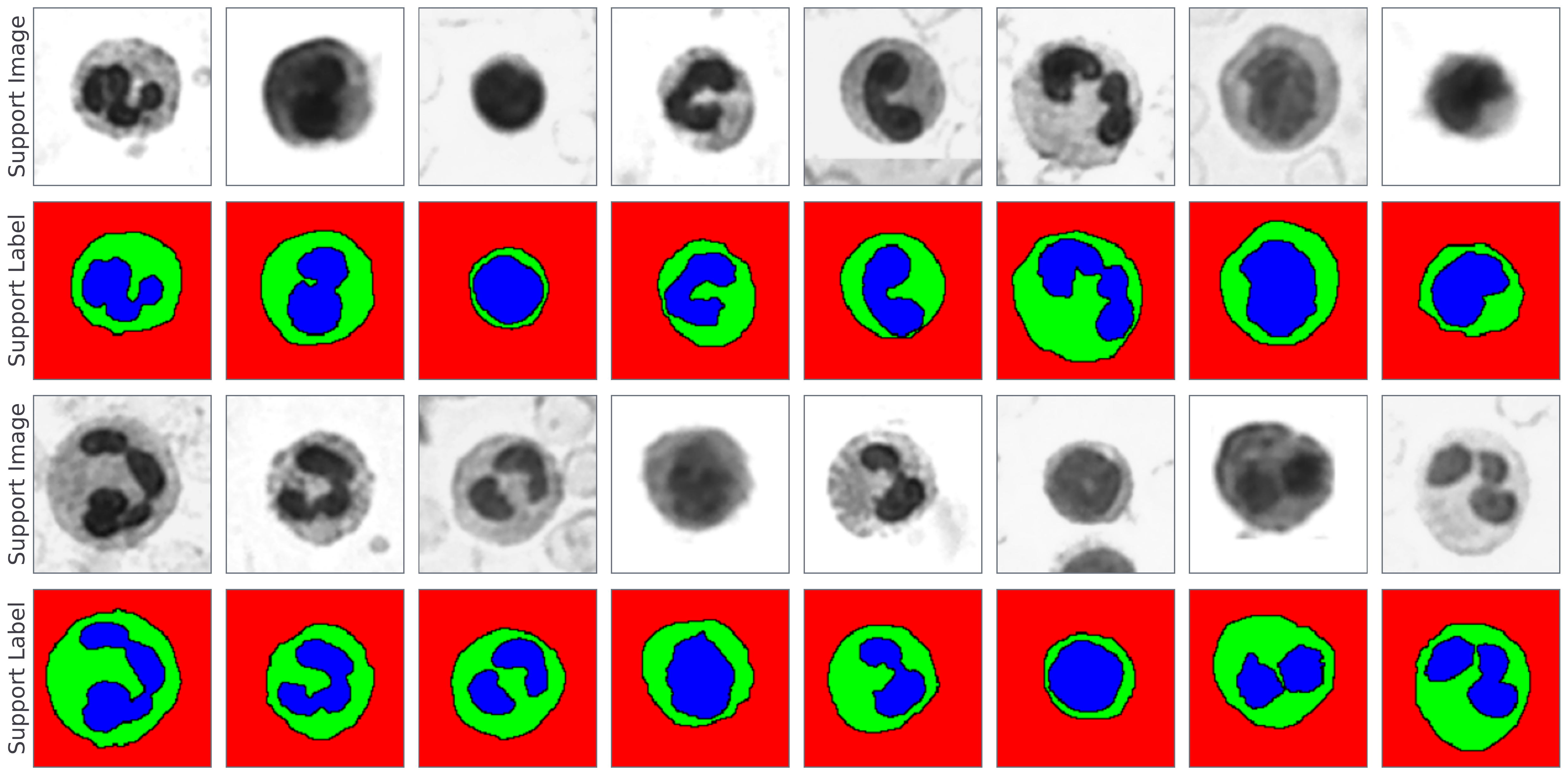}
\end{subfigure}
\begin{subfigure}{\textwidth}
\caption{\textbf{Predictions - Multi Label}}
\includegraphics[width=\textwidth]{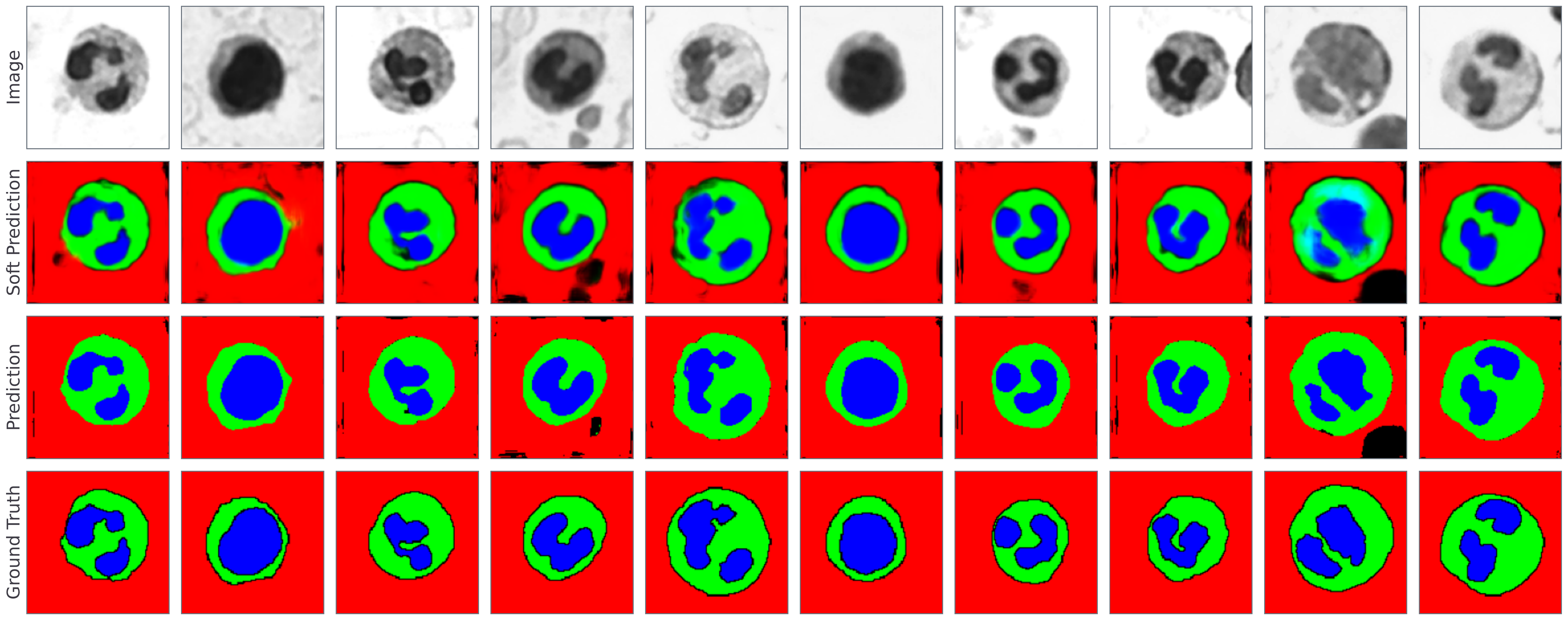}
\end{subfigure}
\caption{Visualization of support set examples (a) and predictions (b) for the WBC task with multiple labels being predicted independently. Each label is encoded using a RGB channel (Red=backgroud, Green=Cytoplasm, Blue=Nuclues), we only see some mild nucleus-cytoplasm overlaps in cyan for one example.}
\label{fig:wbc_viz_05}
\end{figure*}

\begin{figure*}[!h]
\centering
\begin{subfigure}{0.8\textwidth}
\caption{\textbf{Predictions - Example A}}
\includegraphics[width=\textwidth]{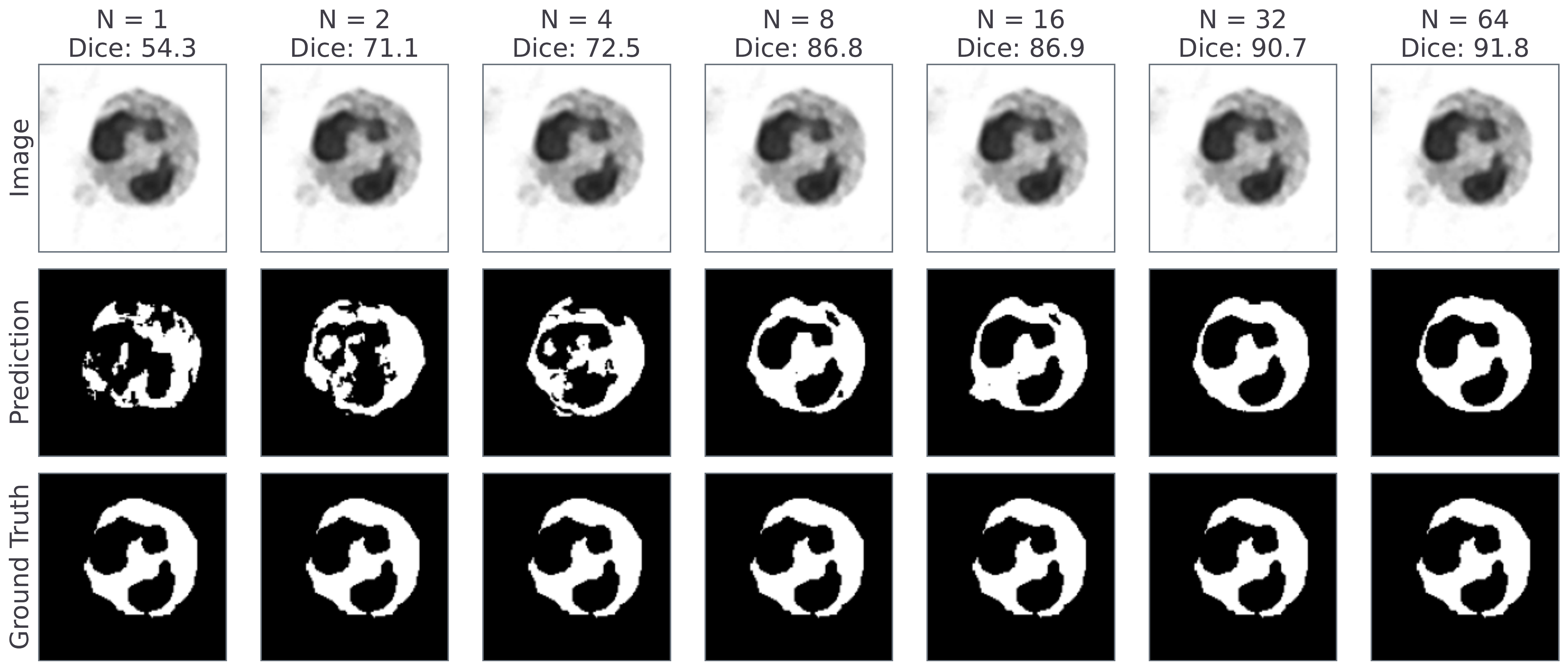}
\end{subfigure}
\begin{subfigure}{0.8\textwidth}
\caption{\textbf{Predictions - Example B}}
\includegraphics[width=\textwidth]{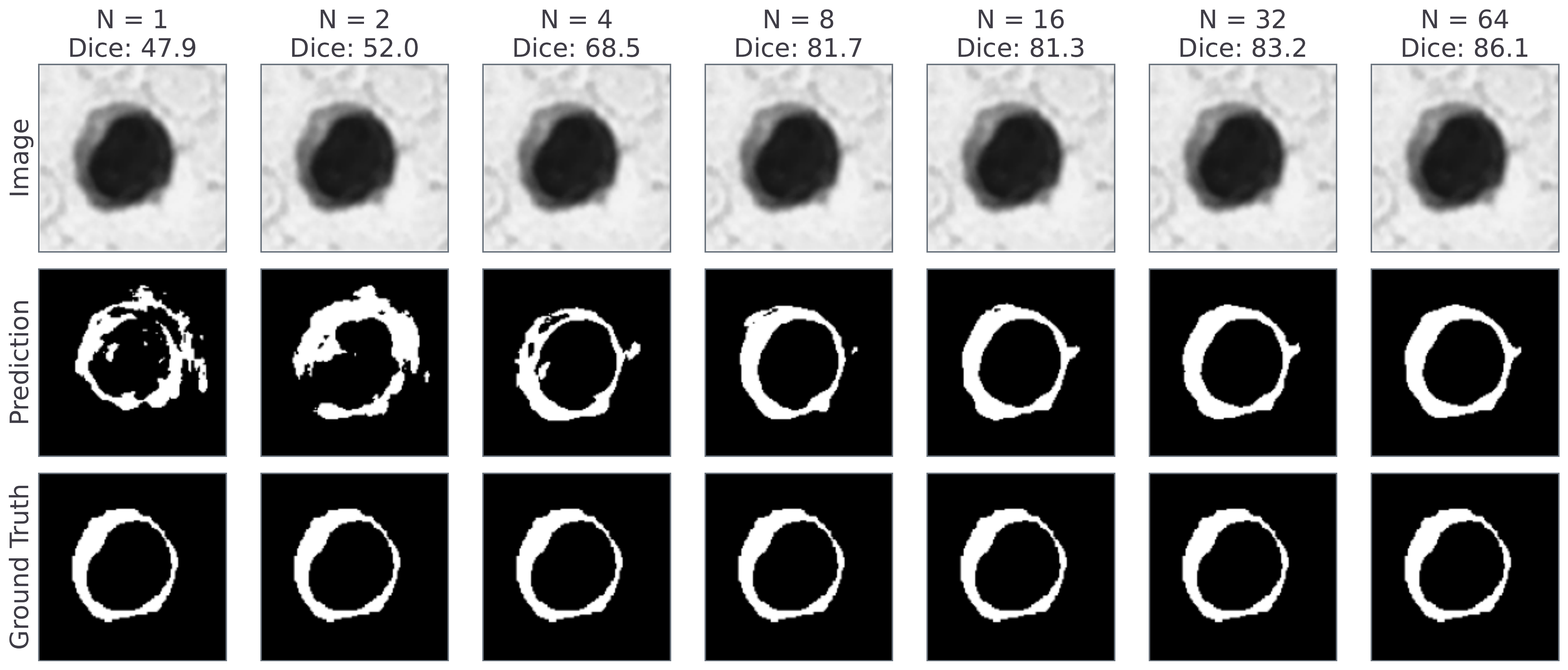}
\end{subfigure}
\begin{subfigure}{0.8\textwidth}
\caption{\textbf{Predictions - Example C}}
\includegraphics[width=\textwidth]{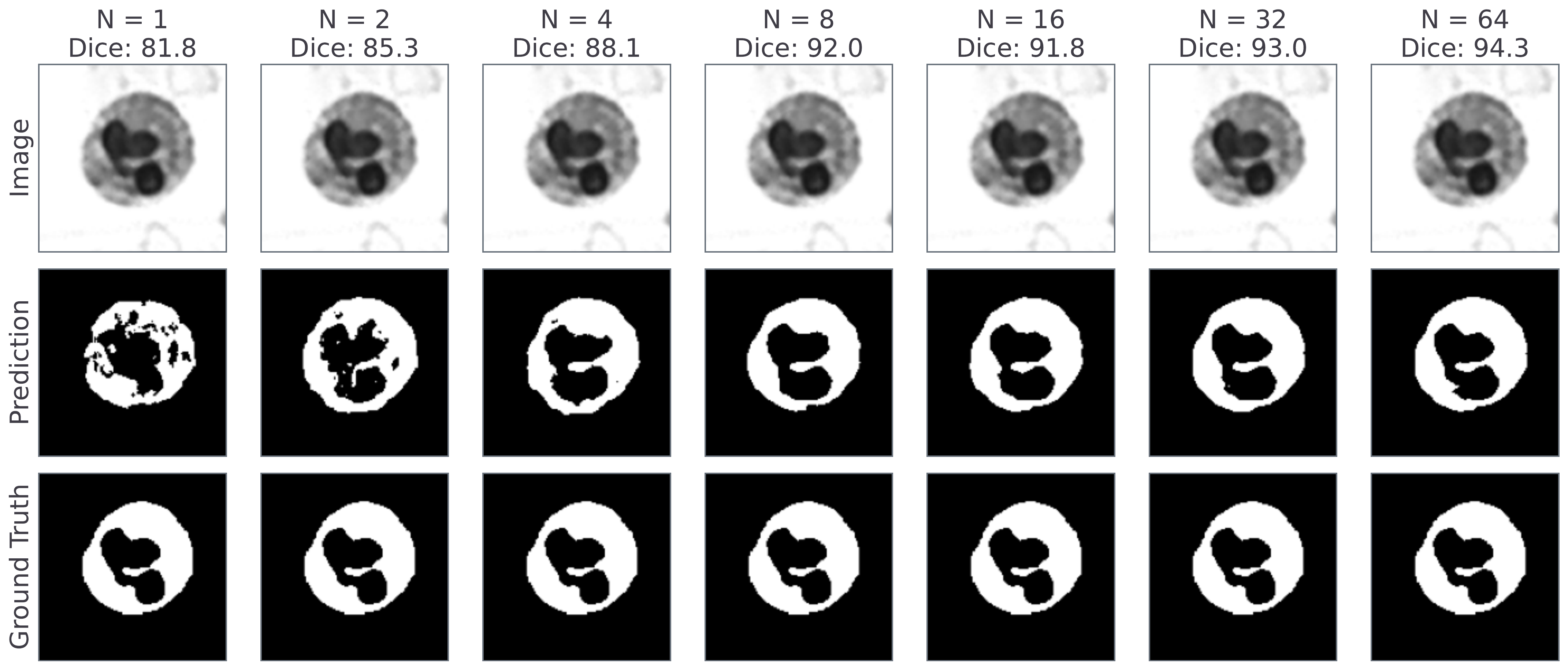}
\end{subfigure}
\caption{Visualization of predictions for the WBC Cytoplasm task with varying number of support set examples $N$. Larger support sets lead to better segmentation masks.}
\label{fig:wbc_viz_04}
\end{figure*}

\begin{figure*}[!h]
\centering
\begin{subfigure}{0.7\textwidth}
\caption{\textbf{Predictions - Example A}}
\includegraphics[width=\textwidth]{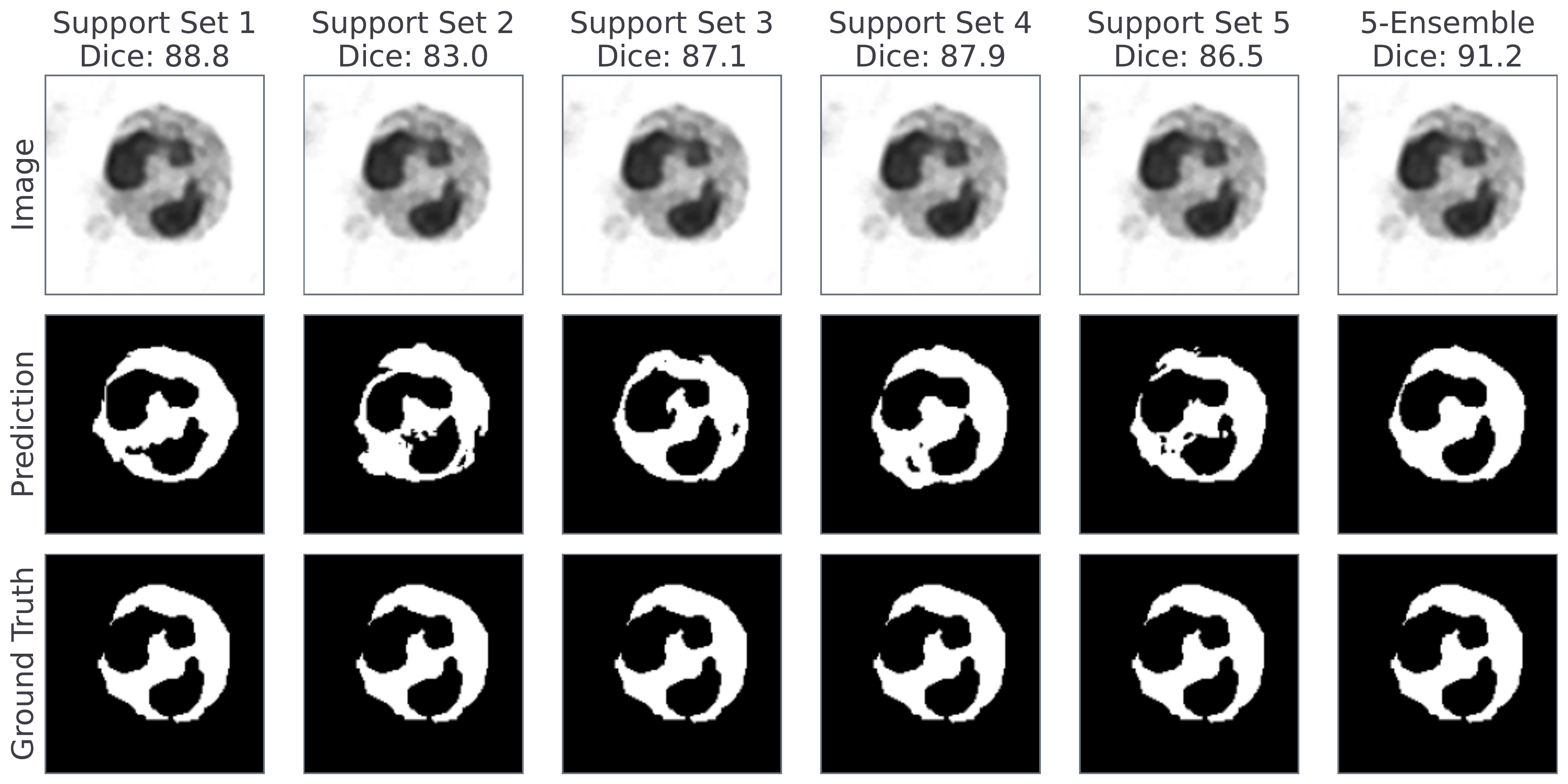}
\end{subfigure}
\begin{subfigure}{0.7\textwidth}
\caption{\textbf{Predictions - Example B}}
\includegraphics[width=\textwidth]{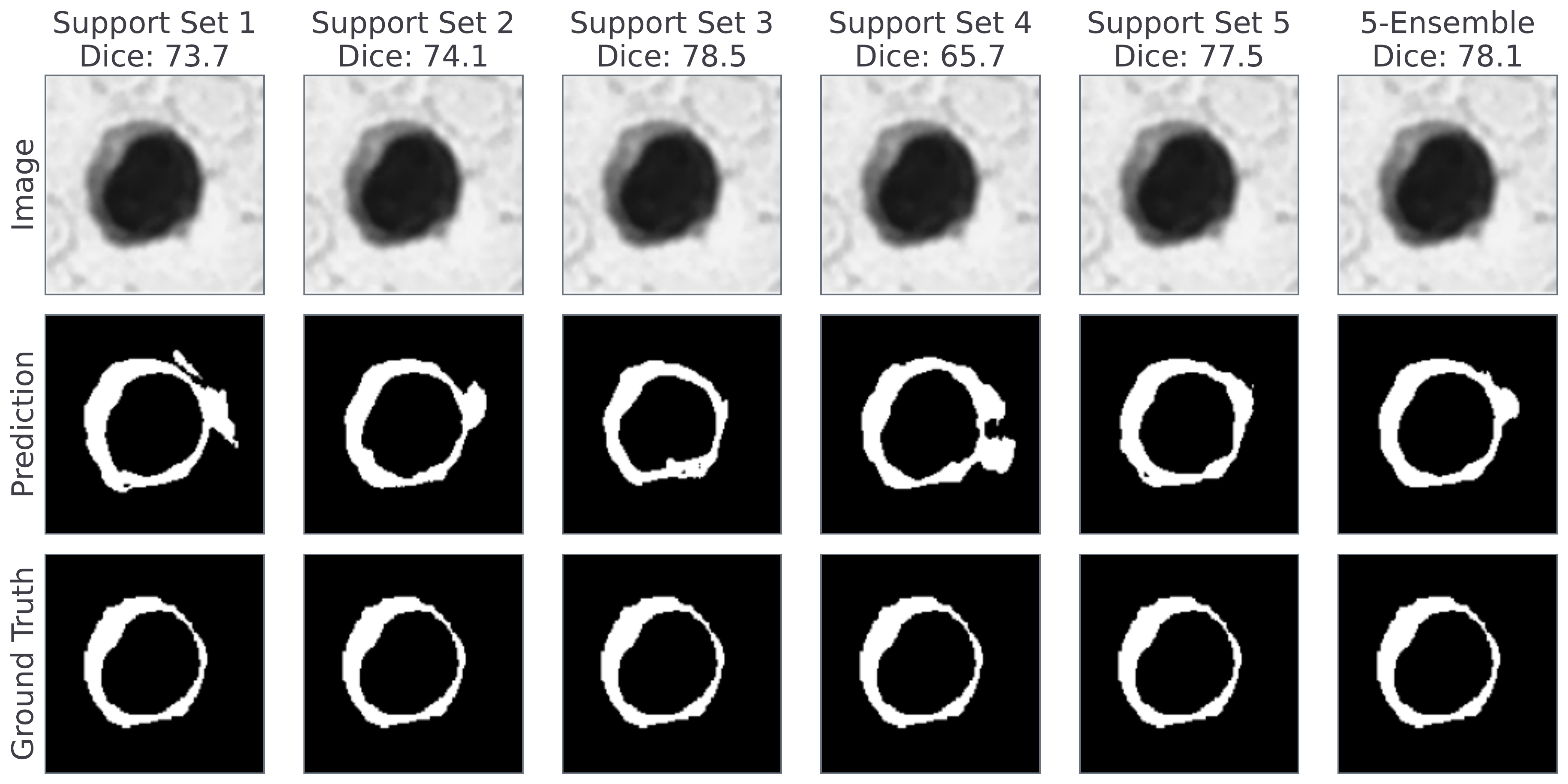}
\end{subfigure}
\begin{subfigure}{0.7\textwidth}
\caption{\textbf{Predictions - Example C}}
\includegraphics[width=\textwidth]{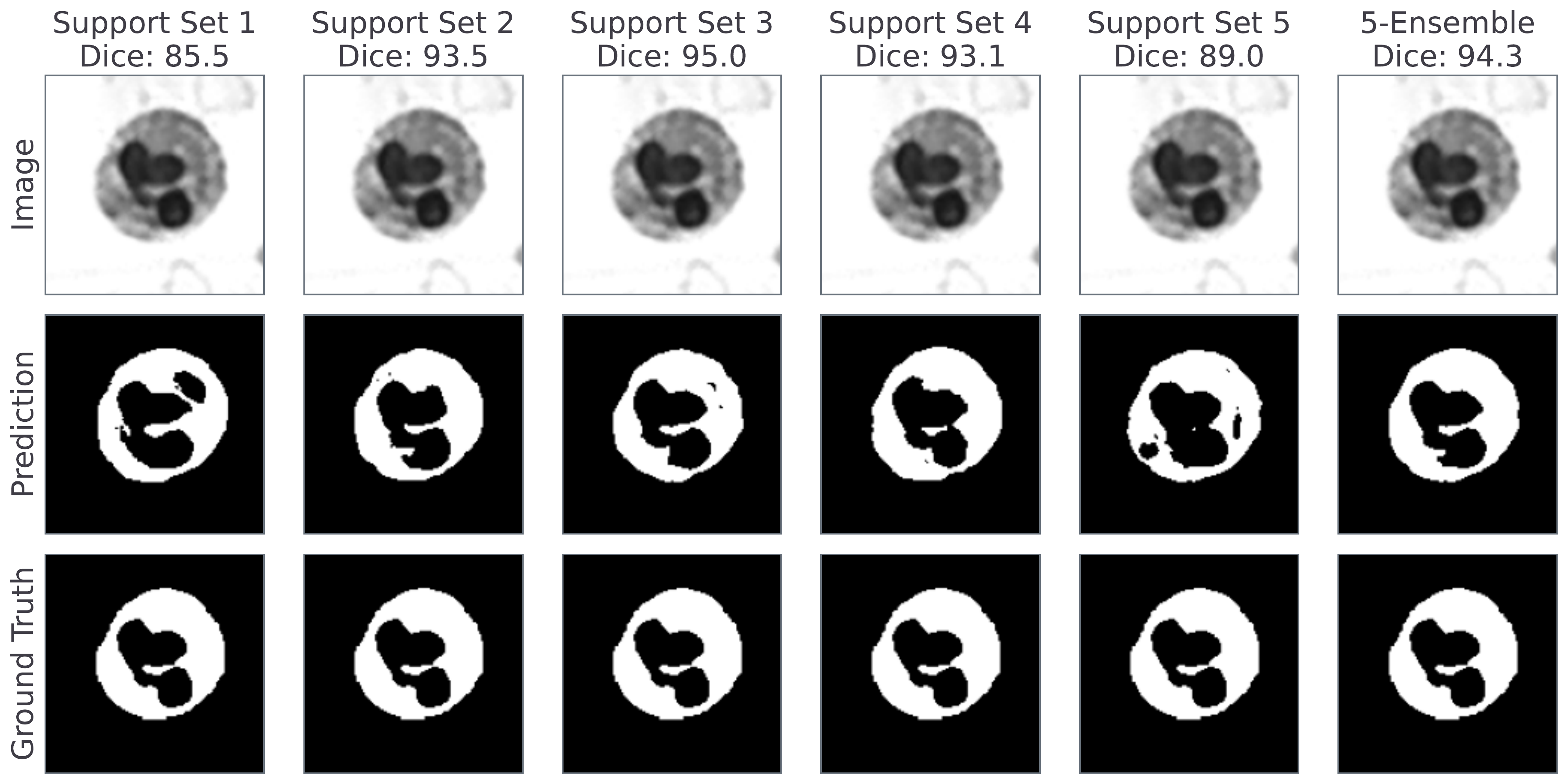}
\end{subfigure}
\caption{Visualization of predictions for the WBC Cytoplasm task with various choices of support set ($N=8$) as well as the ensembled prediction (last column). Ensembling reduces the variance of predictions.}
\label{fig:wbc_viz_06}
\end{figure*}

\begin{itemize}
  \item Figure~\ref{fig:wbc_viz_01} presents support set examples for the WBC Cytoplasm label as well as held-out predictions, showing that \method{} closely matches the ground truth.
  \item Figure~\ref{fig:wbc_viz_02} shows how \method{} is equivariant with respect to the support set labels. Given the same images as in Figure~\ref{fig:wbc_viz_01} but different labels, \method{} adapts its predictions to the nucleus label.
  \item Figure~\ref{fig:wbc_viz_03} showcases \method{}'s invariance to image transformations. Using the same images and label examples from Figure~\ref{fig:wbc_viz_01}, we invert the image data (i.e. $1-x$) for both the query and support set images. \method{} correctly segments the label regardless of the image transformation.
  \item Figure~\ref{fig:wbc_viz_05} shows that while \method{} is trained on binary segmentation tasks, it can adequately perform  multi-label segmentation. To produce multi-label predictions, we treat each label independently, and then combine the predictions for each label using a softmax operation.
  \item Figure~\ref{fig:wbc_viz_04} shows the effect of the support set size $N$ in the prediction results. We observe that segmentation mask quality substantially improves as we increase the number of support set image-label pairs.
  \item Figure~\ref{fig:wbc_viz_06} shows prediction variability for predictions performed with support size $N=8$ along with an ensembled prediction.
\end{itemize}

\end{document}